\documentclass[iicol, sn-mathphys]{sn-jnl}

\usepackage{pythonhighlight}
\usepackage{amssymb,amsmath,amsthm}
\usepackage{csquotes}
\usepackage{xurl}

\jyear{2021}

\newtheorem{theorem}{Theorem}
\newtheorem{lemma}{Lemma}[theorem]
\newtheorem{definition}{Definition}[theorem]

\begin{document}

\title[Article Title]{RELAX: Representation Learning Explainability}

\author*[1]{\fnm{Kristoffer K.} \sur{Wickstrøm}}\email{kristoffer.k.wickstrom@uit.no}

\author[1]{\fnm{Daniel J.} \sur{Trosten}}\email{daniel.j.trosten@uit.no}

\author[1]{\fnm{Sigurd} \sur{Løkse}}\email{sigurd.lokse@uit.no}

\author[1]{\fnm{Ahcène} \sur{Boubekki}}\email{ahcene.boubekki@uit.no}

\author[1]{\fnm{Karl Øyvind} \sur{Mikalsen}}\email{karl.o.mikalsen@uit.no}

\author[1]{\fnm{Michael C.} \sur{Kampffmeyer}}\email{michael.c.kampffmeyer@uit.no}

\author[1]{\fnm{Robert} \sur{Jenssen}}\email{robert.jenssen@uit.no}

\affil*[1]{\orgdiv{Department of Physics and Technology}, \orgname{UiT The Arctic University of Norway}, \orgaddress{\street{Hansine Hansens veg 18}, \city{Tromsø}, \postcode{9019}, \state{Troms}, \country{Norway}}}

\abstract{Despite the significant improvements that self-supervised representation learning has led to when learning from unlabeled data, no methods have been developed that explain what influences the learned representation. We address this need through our proposed approach, RELAX, which is the first approach for attribution-based explanations of representations. Our approach can also model the uncertainty in its explanations, which is essential to produce trustworthy explanations. RELAX explains representations by measuring similarities in the representation space between an input and masked out versions of itself, providing intuitive explanations and significantly outperforming the gradient-based baselines. We provide theoretical interpretations of RELAX and conduct a novel analysis of feature extractors trained using supervised and unsupervised learning, providing insights into different learning strategies. Moreover, we conduct a user study to assess how well the proposed approach aligns with human intuition and show that the proposed method outperforms the baselines in both the quantitative and human evaluation studies. Finally, we illustrate the usability of RELAX in several use cases and highlight that incorporating uncertainty can be essential for providing faithful explanations, taking a crucial step towards explaining representations.}




\maketitle

\section{Introduction}\label{intro}

Interpretability is of vital importance for designing trustworthy and transparent deep learning-based systems \citep{Pedreschi2019, sanaMLH}, and the field of explainable artificial intelligence (XAI) has made great improvements over the last couple of years \citep{antoran2021getting, Schulz2020Restricting}. However, there exists no methods for attribution-based explanations of \emph{representations}, despite the tremendous advances in representation learning using e.g self-supervised learning \citep{simclr, swav, moco}. This lack of explainability makes representation learning less trustworthy and dependable, and there is therefore a need for representation learning explainability. To be able to explain learned representations would provide crucial information in several use-cases. For instance, a typical clustering approach is applying K-means to the representation produced by a feature extractor trained on unlabeled data \citep{completer, cdimc, kmeansfriend}, but there is no method for investigating which features are characteristic for the members of a cluster.

\begin{figure}[t!]
    \centering
    \includegraphics[width=\linewidth]{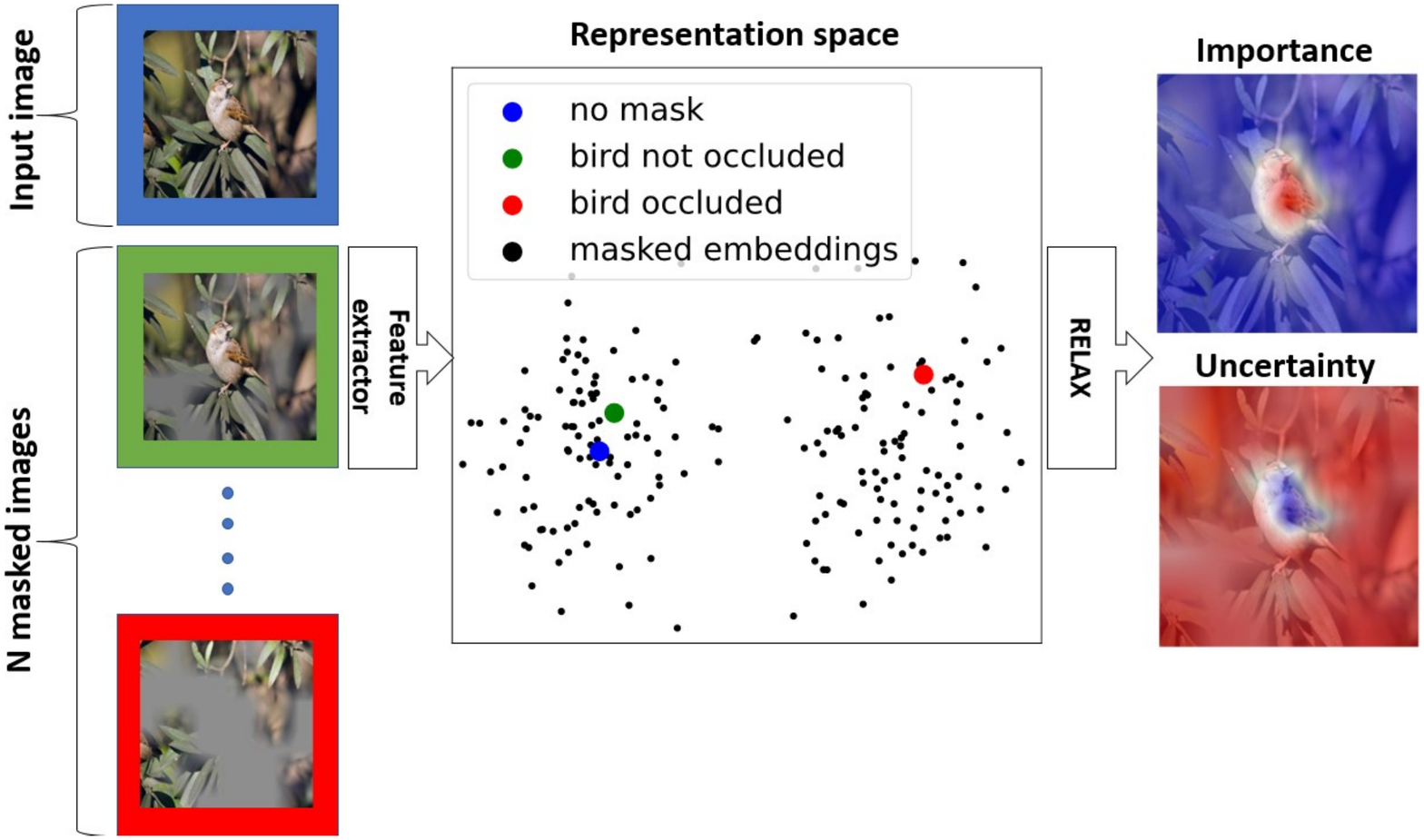}
    \caption{Conceptual illustration of RELAX. An image is passed through an encoder that produces a new vector representation of the image. Similarly, masked images are embedded in the same latent space. Input feature importance is estimated by measuring the similarity between the representation of the unmasked input with the representations of numerous masked inputs.}
    \label{fig:IllustrationOfMethod}
\end{figure}

Representation learning explainability would also allow for a new approach for evaluating representation learning frameworks. Representation learning frameworks are typically evaluated by training simple classifiers on the representation produced by the feature extractor or through a downstream task \citep{simclr, moco, swav}. However, such approaches provide only limited information about the features used by the models, and might ignore important distinctions between them. For instance, a similar accuracy on some downstream task does not necessarily equate to the representations being based on the same features. This highlights the need for an explanatory framework for representations, as many of the current evaluation methods are not sufficient for illuminating differences in the what features are used by different feature extractors.

\begin{figure*}[ht]
    \centering
    \includegraphics[width=\linewidth]{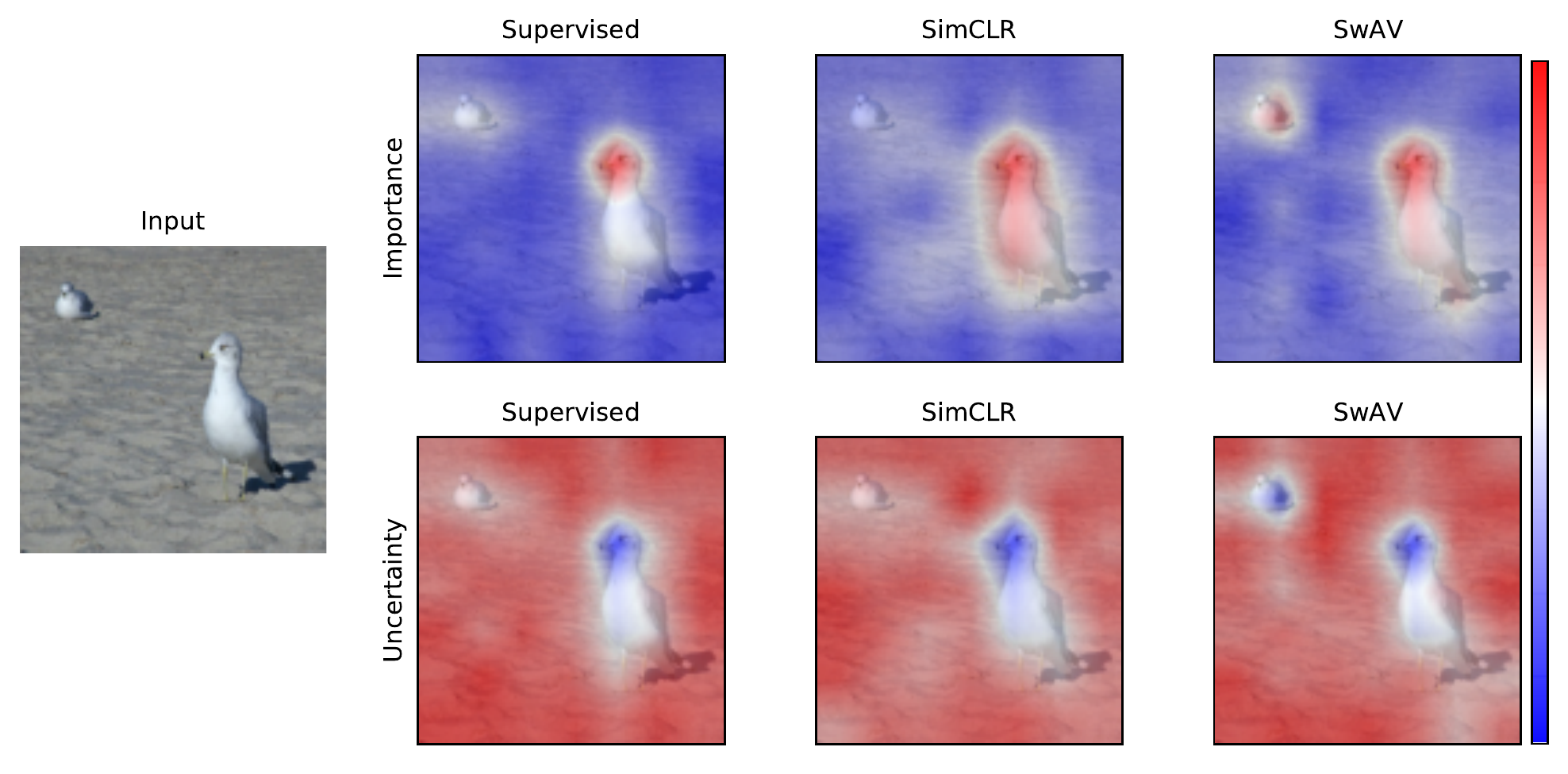}
    \caption{The figure shows the RELAX explanation and its uncertainty for the representation of the leftmost image for three widely used feature extractors. The first row displays the explanations for the representation and the second row shows the uncertainty associated with the different explanations. Red indicates high values and blue indicates low values. In this example, two objects are present in the image, one bird prominently displayed in the foreground, and another more inconspicuous bird in the background. The plots show that all models emphasize the bird in the foreground with low uncertainty. On the other hand, there is more disagreement on how much emphasis to put on bird in the background, also with a differing degree of uncertainty. The example illustrates that different feature extractors utilize different features in the representation of the image, and with different amounts of uncertainty. The image is taken from VOC \citep{Everingham2009}.}
    \label{fig:FirstPage}
\end{figure*}

However, any explanatory framework can make over or under-confident explanations. Hence, uncertainty is a key component for designing trustworthy models, since trusting an explanation without knowing the uncertainty of the explanation might lead to an unjustified trust in the model. A recent survey where clinicians were asked what was necessary for making trustworthy models, found that explainability alone was not enough and that uncertainty was also of high importance \citep{sanaMLH}. Our experiments show that uncertainty can be used to increase the faithfulness of explanations, by removing uncertain parts of an explanations. Nevertheless, little work has been done on uncertainty in explanations of representations.

In this work, we present the first framework for explaining representations, entitled REpresentation LeArning eXplainability (RELAX), which is also equipped with uncertainty quantification with respect to its own explanations. The framework is illustrated in Figure \ref{fig:IllustrationOfMethod}. RELAX measures the change in the representation of an image when compared with masked versions of itself. The core idea is that when informative parts of the input are masked out, the representation should change significantly. When averaging over numerous masks, RELAX reveals the important regions of the input. RELAX is an intuitive and highly versatile framework that can explain any representation, given a suitable similarity function and masking strategy. To provide insight into the geometrical properties of RELAX, we show that the importance of a pixel can be seen as the result of a scoring function based on an inner product between the input and the mean of the masked representations in the representation space. Figure \ref{fig:FirstPage} shows an example where RELAX is used to investigate the explanations and the corresponding uncertainties for a selection of widely used feature extraction models, which demonstrate that RELAX is a versatile framework for highlighting the emphasis that feature extractors put on pixels and regions in the input (top row).

Our contributions are:
\begin{itemize}
    \item RELAX, a novel framework for explaining representations that also quantifies its uncertainty.
    \item A threshold approach called U-RELAX that removes uncertain parts of an explanation and increases the faithfulness of the explanations.
    \item A theoretical analysis of the framework and an estimation of the number of masks needed to obtain a given level of confidence.
    \item A comprehensive experimental section that compares widely used supervised and self-supervised feature extraction models and evaluates a number of hyperparameters.
    \item A user study that examines how well the explanations align with human evaluation.
    \item Two use cases for RELAX. First, RELAX enables explainability in state-of-the-art incomplete multi-view clustering. This illustrates the usability of RELAX in recent cutting-edge research. Second, RELAX allows for explanation of classic computer vision techniques such as Histogram of Oriented Gradients (HOG). This demonstrates that RELAX is a flexible framework, which is capable of explaining representations produced by any method, not just those produced by deep neural networks.
\end{itemize}

Code for RELAX is available at \url{https://github.com/Wickstrom/RELAX}.

\section{Related Work}\label{relwork}

In this section, we present the previous works that are most closely related to our work. The focus will be on attribution-based explanations where each input feature is assigned an importance. Therefore, we will not consider other explainability methods such as example-based explanations \citep{influence, KarBarBalVal20} or global explanations \citep{45507}.

\textbf{Occlusion-based explainability}. There exist a number of occlusion-based explainability methods. Systematically occluding an image with a gray rectangle and then measuring the change in activations could be used to provide coarse explanations for CNNs \citep{ZeilerFergus}. A more sophisticated occlusion approach can improve explanations, in which smooth masks are generated and accumulated to produce explanations for the prediction of a model \citep{Petsiuk2018rise}. A slightly different approach is meaningful perturbations, where a spatial perturbation mask that maximally affects the model's output is optimized \citep{8237633}. A follow up work proposed extremal perturbations, where a perturbation can be considered extremal if it has maximal effect on the network's output among all perturbation of a given, fixed area \citep{Fong2019ICCV}. On a different note, an information theoretic approach to XAI has been proposed, where noise is injected in order to measure the information in different regions of the input \citep{Schulz2020Restricting}. Similarly, \cite{kolek2021ratedistortion} introduced a rate-distortion perspective to explainability. Note that none of these methods are capable of providing explanations for representations.

\textbf{Explaining representations}. Attribution-based explainability methods are extensively used to explain specific sample predictions \citep{bachplos, Petsiuk2018rise, Schulz2020Restricting}. However, to the best of our knowledge, no attribution-based explainability method exists for explaining representations. 
While initial attempts have been made to explain representations such as the Concept Activation Vectors \citep{tcav}, which uses directional derivatives to quantify the model prediction's sensitivity, these explanations only relate the representations to high-level concepts and require label information. Similarly, network dissection has been proposed to interpret representations \citep{netdissect2017}, but requires predefined concepts and label information without indicating the importance of individual pixels. A different direction is designing models that have the capability to explain their own decisions built into the system \citep{protoPnet, robustXAI}. Two drawbacks of such an approach is that it might lead to models with weaker performance and does not explain representations. Another approach maps semantic concepts to vectorial embedding \citep{Fong2018CVPR}. However, this requires segmentation masks that are not available in the unsupervised setting. Representations have also been investigated from learnability and describability perspectives \citep{laina20quantifying}, but this was achieved through human-annotators that are typically not available. Lastly, the inspectability of deep representations have been investigated through an information bottleneck approach \citep{Losch2021}, but with a focus on segmentation and predefined concepts.

\textbf{Uncertainty in explainability}. Modeling uncertainty in explainability is a rapidly evolving research topic that is receiving an increasing amount of attention. One of the earliest works proposed to use Monte Carlo Dropout \citep{MCdrop} in order to estimate the uncertainty in gradient-based explanations \citep{8516998, WICKSTROM2020101619}, which was later followed by a similar approach that was based on Layer-wise Relevance Propagation \citep{Kiril}. Uncertainties that are inherent in the widely used LIME method \citep{lime} have been explored \citep{LimeUc}. Also, ensemble-based approaches, where uncertainty estimates are obtained by taking the standard deviation across the ensemble, have also been proposed \citep{Wick2}. Recently, Counterfactual Latent Uncertainty Explanations (CLUE) was presented \citep{antoran2021getting}, where uncertainty estimates from probabilistic models can be interpreted. Nevertheless, none of these approaches were designed for quantifying the uncertainty in explanations of representations, as they either require label information or are computationally impractical.

\section{Representation Learning Explainability}
We present RELAX, our proposed method for explaining representations, equipped with uncertainty quantification. Furthermore, we leverage RELAX's ability to quantify uncertainty and introduce as a new concept a method for filtering out uncertain parts of the explanations, which we entitle U-RELAX. This is important, as uncertain explanations might give an unwarranted trust in the model. Our framework is inspired by RISE \citep{Petsiuk2018rise}. However, RISE was designed for explaining predictions and is not transferable for explaining representations or quantifying uncertainty. Note that the proofs of the theorems in this section are given in Appendix \ref{proofs}.

\subsection{RELAX}\label{sec:relax}
The central idea of RELAX is that when informative parts are masked out, the representation should change significantly. Let $\mathbf{X}\in \mathbb{R}^{H\times W}$ represent an image\footnote{To enhance readability, we do not include image channels, but this can be easily included by letting the masks span the channel dimension.} consisting of $H\times W$ pixels, and $f$ denote a feature extractor that transforms an image into a representation $\mathbf{h} = f(\mathbf{X}) \in \mathbb{R}^{D}$. To mask out regions of the input, we apply a stochastic mask $\mathbf{M}\in [0, 1]^{H\times W}$, where each element $M_{ij}$ is drawn from some distribution.

The stochastic variable $\bar{\mathbf{h}} = f(\mathbf{X} \odot \mathbf{M})$, where $\odot$ denotes element-wise multiplication, is a representation of a masked version of $\mathbf{X}$. Moreover, we let $s(\mathbf{h}, \bar{\mathbf{h}})$ represent a similarity measure between the unmasked and the masked representation. Intuitively, $\mathbf{h}$ and $\bar{\mathbf{h}}$ should be similar if $\mathbf{M}$ masks \emph{non-informative} parts of $\mathbf{X}$. Conversely, if \emph{informative} parts are masked out, the similarity between the two representations should be low.

Motivated by this intuition, we define the importance $R_{ij}$ of pixel $(i,j)$ as:

\begin{equation}\label{eq:rel1}
    R_{ij} = \mathrm{E}_{\mathbf{M}}\big[s(\mathbf{h}, \bar{\mathbf{h}})M_{ij}\big].
\end{equation}
Equation \eqref{eq:rel1} is core to our framework as it computes the importance of a pixel $(i, j)$ as a weighted similarity score for masked versions of a given image. However, integrating over the entire support of $\mathbf{M}$ is not computationally feasible. Therefore, we approximate the expectation in Equation \eqref{eq:rel1} by sampling $N$ masks and computing the sample mean:

\begin{equation}\label{eq:rel2}
    \bar{R}_{ij} = \frac{1}{N}\sum\limits_{n=1}^N s(\mathbf{h}, \bar{\mathbf{h}}_n)M_{ij}(n).
\end{equation}
Here, $\bar{\mathbf{h}}_n$ is the representation of the image masked with mask $n$, and 
$M_{ij}(n)$ the value of element $(i, j)$ for mask $n$. The explanations of RELAX are computed through Equation \eqref{eq:rel2}, and an illustration of RELAX is given in Figure \ref{fig:IllustrationOfMethod}. As a similarity measure we use the cosine similarity

\begin{equation}\label{eq:sim}
    s(\mathbf{h}, \bar{\mathbf{h}}) = \frac{\langle \mathbf{h}, \bar{\mathbf{h}}\rangle }{\lVert\mathbf{h}\rVert \lVert\bar{\mathbf{h}}\rVert},
\end{equation}
where $\lVert\cdot\rVert$ denotes the Euclidean norm of a vector. There are several motivations for this choice. First, \cite{Liuetal21} argued that angular information preserves the essential semantics in neural networks, in contrast to magnitude information. Since the cosine kernel normalizes the representation, essentially discarding magnitude information, such a similarity measure would be suited to capture key information encoded in the representations. Second, the cosine kernel does not rely on hyperparameters that must be selected, which may be beneficial in an unsupervised setting where we cannot do cross validation. Third, a large portion of feature extractors trained using self-supervised learning use the cosine kernel in their loss function \citep{simclr, Simsiam}. Therefore, it is the natural choice for measuring similarities in their latent space. However, based on the two first points, the cosine kernel is still suitable for models trained without the cosine kernel. Lastly, other alternatives for the kernel functions, such as the radial basis function or polynomial kernel, requires careful tuning of hyperparameters. We consider an investigation of such alternatives and their hyperparameters as a direction for future research.

First, a large portion of feature extractors trained using self-supervised learning use the cosine kernel in their loss function \citep{simclr, Simsiam}. Therefore, it is the natural choice for measuring similarities in their latent space.

Note that we recognize that the masking strategy can introduce a shift in the distribution of pixel intensities. However, in our experiments, we observed that this potential shift did not impact the explanations. An experiment where the distribution is approximately preserved is included in Appendix \ref{distshift}.

\noindent \textbf{Masking distribution}. There are several ways to sample the masks in Equation \eqref{eq:rel2}, for instance by letting each $M_{ij}(n)$ be iid. Bernoulli. However, sampling masks with the same size as the input results in a massive sample space, and simultaneously makes it challenging to create smooth masks that cover different portions of the image \footnote{See Appendix \ref{maskinstrat} for evaluation of masking strategies.}.

To avoid these problems, we generate masks as suggested by \cite{Petsiuk2018rise}. Binary masks of smaller size than the input image are generated, where each element of these smaller masks is sampled from a Bernoulli distribution with probability $p$. These masks are then upsampled using bilinear interpolation to the same size as the image. The distribution for $M_{ij}$ is then a continuous distribution between 0 and 1. Specifically: we sample $N$ binary masks, each with size $h\times w$, where $h<H$ and $w<W$. We upsample these masks to size $(h+1)C_H \times (w+1)C_W$, where $C_H\times C_W=\lfloor H/h\rfloor\times\lfloor W/w\rfloor$ is the size of the cell in the upsampled masks. Lastly, we crop the final masks of size $H\times W$ randomly from the $(h+1)C_H \times (w+1)C_W$ masks.

\textbf{Number of masks required}. In order to minimize the computational cost of RELAX, we derive the following lower bound on the number of masks required for a certain estimation error.

\begin{theorem}\label{teo:masks}
    Suppose $s(\cdot, \cdot)$ is bounded in $(0, 1)$.\footnote{This holds for the cosine similarity, since the representations considered are assumed to be ReLU outputs (non-negative).} Then, for any $\delta \in (0, 1)$ and $t > 0$, if $N$ in Equation \eqref{eq:rel2}, satisfies:
    \begin{equation}\label{eq:maskeq}
        N \ge -\frac{\ln (\delta/2)}{2t^2},
    \end{equation}
    we have $\mathrm{P}(\lvert\bar{R}_{ij} - R_{ij}\rvert\ge t)\le \delta$.
\end{theorem}

Theorem \ref{teo:masks} states that if $N$ satisfies Equation \eqref{eq:maskeq}, we are able to estimate $R_{ij}$ to an absolute error of less than $t$ with probability at least $1-\delta$. See Appendix \ref{proofs} for proof and verification of bound. In all of our experiments, we generate 3000 masks, which ensures an estimation error below 0.01 with a probability of 0.99.

\textbf{RELAX from a kernel perspective}. To provide insights into the geometrical properties of RELAX, we present a kernel viewpoint of Equation \eqref{eq:rel2}.
\begin{theorem} \label{thm:rkhs}
    Suppose the similarity function $s(\cdot, \cdot)$ is a valid Mercer kernel \citep{mercer1909functions}. The importance $\bar{R}_{ij}$ then acts as a linear scoring function between $\mathbf h$, and the weighted mean of $\bar{\mathbf h}_1, \dots, \bar{\mathbf h}_N$, in the reproducing kernel Hilbert space (RKHS) induced by $s(\cdot, \cdot)$. That is:
    \begin{align}
        \bar R_{ij} = \langle \phi(\mathbf{h}), \frac{1}{N}\sum_{n=1}^N\phi(\bar{\mathbf{h}}_n) M_{ij}(n)\rangle_{\mathcal H},
    \end{align}
    where $\phi: \mathbb R^d \to \mathcal H $ is the mapping to the RKHS, $\mathcal H$, induced by the kernel $s(\cdot, \cdot)$, and $\langle \cdot, \cdot \rangle_{\mathcal H}$ is the inner product on $\mathcal H$.
\end{theorem}

Theorem \ref{thm:rkhs} provides interesting insight, as many scoring functions are based on inner-products, e.g. between points of interest and class-conditional means (e.g., Fisher discriminant analysis, Bayes classifier under Gaussian distributions with equal covariance structure). This means that even though RELAX is a novel approach, it is founded in well-known statistical concepts \citep{mccullagh1989generalized}.

Additionally, RELAX has the following interpretation from non-parametric statistics
\begin{theorem}\label{thm:parzen}
    Suppose $s(\cdot, \cdot)$ is a valid Parzen window \citep{Theodoridispatrec}. Then:
    \begin{align}
        \bar R_{ij} \propto p_{ij}(\mathbf h),
    \end{align}
    where $p_{ij}(\cdot)$ is a weighted Parzen density estimate \citep{Parzen1962} of the density of the masked embeddings:
    \begin{align}
        p_{ij}(\cdot) = \frac{1}{\sum_{n'=1}^{N}M_{ij}(n')} \sum\limits_{n=1}^{N} s(\cdot, \bar{\mathbf h}_n) M_{ij}(n).
    \end{align}
\end{theorem}
A high RELAX score is obtained when the unmasked representation $\mathbf h$ is close to mean of masked representations, which aligns well with out intuition for RELAX. 

\subsection{Uncertainty in Explanations}\label{uncertainty}

Trusting an explanation without a notion of uncertainty can lead to an unjustified faith in the model. Therefore, we introduce an approach that allows uncertainty quantification to be incorporated into the RELAX framework. Our intuition for this approach stems from what happens when informative and uninformative parts are masked out. If informative parts are masked out, the similarity score will not only drop, but drop with varying degree. If there is a big variation in the similarity scores for a given pixel, it indicates that the explanation for said pixel is uncertain. Based on this intuition, we propose to estimate the uncertainty in input feature importance as:

\begin{equation}\label{eq:unc1}
    U_{ij} = \mathrm{Var}_{\mathbf{M}} [s(\mathbf{h}, \bar{\mathbf{h}})M_{ij}].
\end{equation}
Again, it is not feasible to integrate over all of $\mathbf{M}$ and $U_{ij}$ is therefore approximated by the sample variance:

\begin{equation}\label{eq:unc2}
    \bar{U}_{ij} = \frac{1}{N}\sum\limits_{n=1}^N (s(\mathbf{h}, \bar{\mathbf{h}}_n) - \bar{R}_{ij})^2 M_{ij}(n).
\end{equation}
Equation \eqref{eq:unc2} estimates the uncertainty of the RELAX-score for pixel $(i, j)$ by measuring the difference between the similarity score and the explanations. To estimate Equation \eqref{eq:unc2}, we must first estimate the importance of a pixel. The uncertainty estimates provided in Equation \eqref{eq:unc2} can be thought of as measuring the spread of pixel importance values in relation to importance estimated using Equation \eqref{eq:rel2}. There are several benefits of our method. First, it requires no labels, which is sometimes used in other uncertainty estimation methods \citep{antoran2021getting}. Secondly, it avoids computationally intense sampling methods, for instance through Monte Carlo sampling \citep{MCBN, MCdrop}. Lastly, the uncertainty estimation can be combined with the computation of Equation \eqref{eq:rel2}, as explained in Section \ref{sec:onepass}.

\begin{figure*}[htb]
    \centering
    \includegraphics[width=0.9\linewidth]{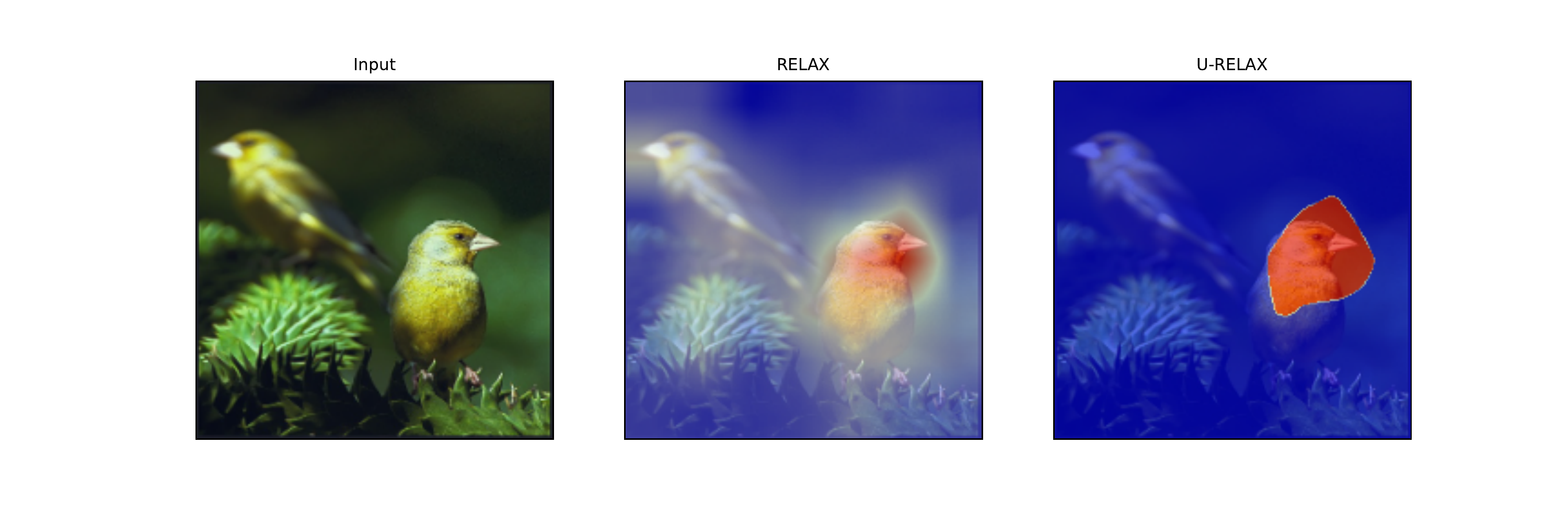}
    \caption{Comparison of RELAX and U-RELAX on an image taken from PASCAL VOC, where red indicates high importance and blue indicates low importance. In this case, the emphasis on the bird in the background is removed as the uncertainty was to high for this part of the explanation.}
    \label{fig:urelax}
\end{figure*}

\subsection{U-RELAX: Uncertainty Filtered Explanations}\label{sec:urelax}
All parts of an explanation do not have the same level of uncertainty associated with it. In such cases, it could be beneficial to remove input features that are indicated as important but also have high uncertainty, while only keeping important input features with low uncertainty. This could increase the faithfulness of an explanation and provide clearer explanations. Therefore, we propose a thresholding approach where explanations with high uncertainty are removed from the explanation. We define our U-RELAX importance score as:
\begin{equation}\label{eq:relbar}
    \bar{R}_{ij}' = \begin{cases}
\bar{R}_{ij},\quad &\text{ if }  \bar{U}_{ij} < \epsilon\\
0,           &\text{ otherwise }
\end{cases},
\end{equation}
where $\epsilon$ is a threshold chosen by the user. Essentially, Equation \eqref{eq:relbar} provides the possibility to only consider explanations of a particular certainty level, depending on $\epsilon$. We propose two ways of choosing epsilon. First as:

\begin{equation}\label{eq:epsilon}
    \epsilon = \frac{\gamma}{HW}\sum_i^H\sum_j^W \bar{U}_{ij},
\end{equation}
that is, the average uncertainty for a particular image, weighted by hyperparameter $\gamma$. This provides a simple and intuitive way of selecting the threshold, which is motivated by only wanting to consider pixels that have high importance and low uncertainty. Alternatively, $\epsilon$ can be computed by taking the median uncertainty for a particular image.

We refer to this uncertainty-filtered version of RELAX as U-RELAX. Figure \ref{fig:urelax} shows an example of the U-RELAX explanation compared with the RELAX explanation. In this case, the emphasis on the bird in the background is removed as the uncertainty was too high for this part of the explanation.

\subsection{One-Pass Version of RELAX } \label{sec:onepass}

Computing Equation \eqref{eq:unc2} requires first computing Equation \eqref{eq:rel2}, since the uncertainty estimation requires an estimate of the importance in order to be computed. This introduces additional computational overhead. We refer to computing Equation \eqref{eq:rel2} followed by Equation \eqref{eq:unc2} as the \textit{two-pass} version of RELAX. To improve computational efficiency, we propose an online version of RELAX where importance and uncertainty is computed simultaneously, which we refer to as the \textit{one-pass} version of RELAX. One-pass RELAX is based on well-known estimators of running mean and variance \citep{onlineEstimator}. Importance is computed as:

\begin{equation}
    \begin{split}
    \bar{R}_{ij}^{(n)} =& \bar{R}_{ij}^{(n-1)}+ \\ & M_{ij}(n)\frac{s(\mathbf{h}, \bar{\mathbf{h}}_n)(n)-\bar{R}_{ij}^{(n-1)}}{W_{ij}(n)},
    \end{split}
\end{equation}
where $\bar{R}_{ij}^{(n)}$ is the importance of pixel $(i, j)$ at mask $n$, and $W_{ij}(n)=\sum_{n'=0}^n M_{ij}(n')$ is the sum of the mask elements $(i,j)$ for the first $n$ masks. Uncertainty is computed as:

\begin{equation}
    \begin{split}
    \bar{U}_{ij}^{(n)} =& \bar{U}_{ij}^{(n-1)}+
    M_{ij}(n)(s(\mathbf{h}, \bar{\mathbf{h}}_n) - \\
    & \bar{R}_{ij}^{(n)})(s(\mathbf{h}, \bar{\mathbf{h}}_n)-\bar{R}_{ij}^{(n-1)}),
    \end{split}
\end{equation}
where $\bar{U}_{ij}^{(n)}$ is the uncertainty in the importance of pixel $(i, j)$ after the $n$th mask. Pseudo-code is shown in Algorithm \ref{alg:relax}. All experiments are carried out using the one-pass version of RELAX. See Appendix \ref{supp:onevstwo} for a comparison of the one-pass versus two-pass version.

\begin{algorithm}\label{alg:1}
\caption{Pytorch-like pseudocode for RELAX.}\label{alg:relax}
\begin{minipage}{\linewidth}
\begin{python}
    # f         - feature extractor
    # X[1,C,H,W]- input image
    # R[H,W]    - importance (init as zeros)
    # U[H,W]    - uncertainty (init as zeros)
    # W[H,W]    - sum of masks (init with 
    #             small positive number)<
    for mask in mask_generator: # [1,1,H,W]
        W += mask
        h, h_mask = f(x), f(x*mask)
        s = cosine_similarity(h, h_mask)
        R_prev = R
        R += m*(s-R)/W
        U += (s-R)*(s-R_prev)*m
    return R, U/(W-1)
\end{python}
\end{minipage}
\end{algorithm}

\section{Evaluation and Baseline}

\subsection{Evaluation of Explanations}\label{sec:scores}

Evaluation is a developing subfield of XAI, and a unifying score is not agreed upon \cite{doshivelez2017rigorous}, even more so for explanations of representations. To evaluate the explanations, we use two of the most widely used explainability evaluation scores, namely localisation and faithfulness \citep{7552539, Petsiuk2018rise, Fong2019ICCV, Schulz2020Restricting}. All scores are computed using the Quantus toolbox \citep{hedstrom2022quantus}.

\textbf{Localisation}. The explanations should put emphasis on input regions corresponding to the objects present in an image. Localisation measures to which degree the explanation agrees with the ground truth location of an object. High performance in localisation indicates that the explanations often align with the bounding boxes or segmentation masks provided by human annotators. We consider three localisation scores, the \textit{pointing game} \citep{Zhang2017}, \textit{top-k intersection}, and \textit{relevance rank accuracy} \citep{ARRAS202214}. The pointing game measures whether the pixel with the highest importance is located within the object location. Top-k intersection considers the binarized version of the top-k most important pixels and measures the intersection with the ground truth mask. Relevance rank accuracy is measured by taking the ratio of high intensity relevances within the ground truth mask. Since RELAX operates in the unsupervised setting we do not have explanations for individual classes. Therefore, the bounding boxes/segmentation masks are collected into one unified bounding box/segmentation mask. This results in unsupervised version of localisation that is suitable for explaining representations.

\textbf{Faithfulness}. Pixels assigned with high importance should be indicative of "true" importance. Faithfulness is typically measures by monitoring the classification accuracy of a classifier as input features are iteratively removed. High faithfulness indicates that the explanation is capable of identifying features that are important for classifying an image correctly. We measure faithfulness using the \textit{monotonicity} score. \cite{Nguyen2020OnQA} proposed to measure monotonicity by computing the correlation of the absolute values of the attributions and the uncertainty in the probability estimation. This will indicate if an explanation is correctly highlighting important features in the input.

\subsection{Representation Explainability Baseline}\label{sec:baseline}

\begin{figure*}[htb]
    \centering
    \includegraphics[width=0.9\linewidth]{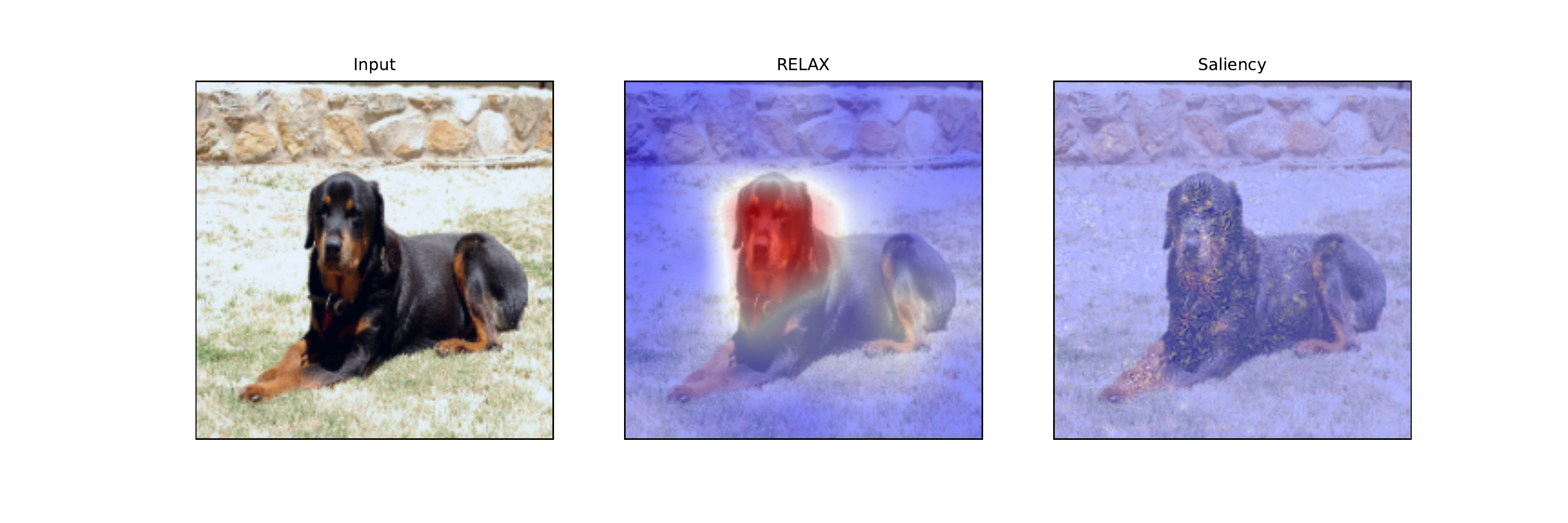}
    \caption{Comparison of RELAX and saliency explanation for an image from PASCAL VOC. The example shows how both explanations focus on the dog, but the saliency explantion is much more erratic and unfocused than the RELAX explanations.}
    \label{fig:goodgrad}
\end{figure*}

\begin{figure*}[htb]
    \centering
    \includegraphics[width=0.9\linewidth]{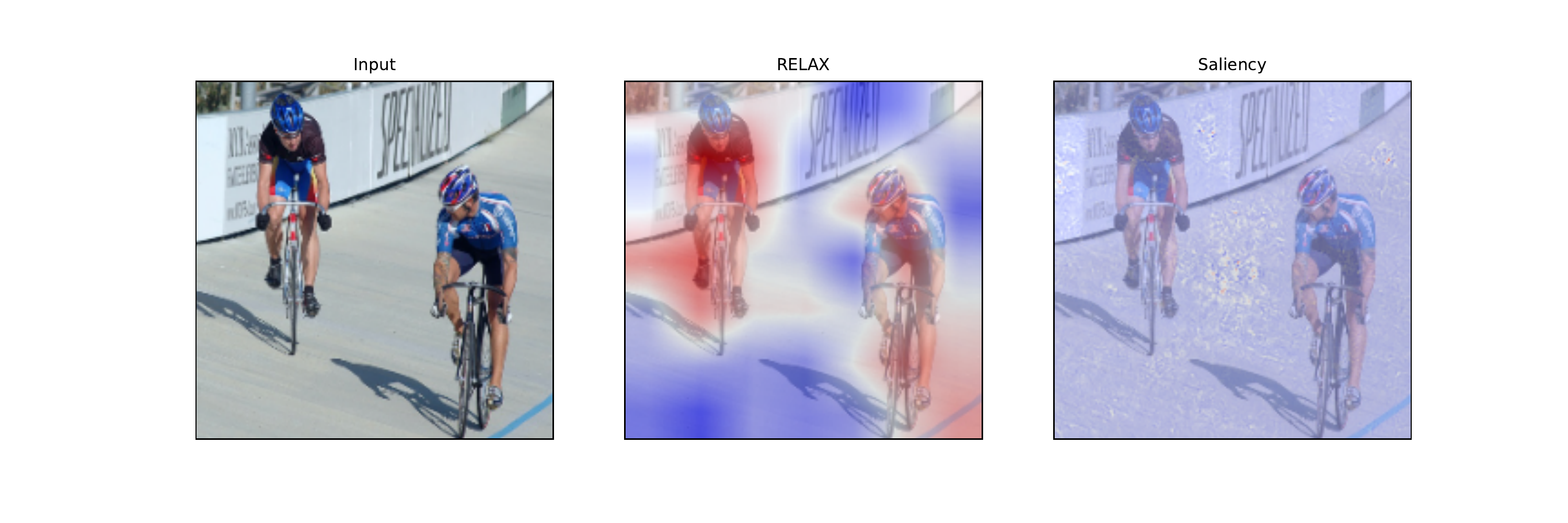}
    \caption{Comparison of RELAX and Saliency explanation for an image from PASCAL VOC. The example shows how RELAX captures information about both objects, while the saliency explanation is focused on the gap in between the two objects.}
    \label{fig:badgrad}
\end{figure*}

While there are are no existing methods that provide attribution-based explanations for representations, it is possible to adopt certain methods to provide such explanations. One of the most common baselines in the field of explainability is saliency explanations \citep{guidebackprop, sanitycheck}, which utilize gradient information to attribute importance. An explanation is obtained by computing the gradient for a prediction with respect to the input. However, it is not trivial to extend such methods for explaining representations. We propose the following for a saliency approach:

\begin{equation}\label{eq:sal1}
    \mathbf{S} = \frac{1}{D}\sum\limits_{d=1}^D \nabla f(\mathbf{X})_d,
\end{equation}
where $D$ is the dimensionality of the representation and $S_{ij}$ is the importance of pixel $(i, j)$ for the given representation. The gradient for each dimension of the representation will give an explanation, and Equation \eqref{eq:sal1} takes the mean across all explanations. This is the most straight-forward and intuitive approach for explaining representations with gradients. It also illustrates the challenges that arise when adopting gradient-based explanations for representation, as some form of agglomeration of the explanations is required. Figure \ref{fig:goodgrad} and Figure \ref{fig:badgrad} shows a qualitative comparison between the RELAX and saliency explanation for a representation of an image. Both Figures illustrate how RELAX provides more intuitive and clear explanations that are able to capture information related to the objects in the image, when compared with the saliency explanation.

Once the saliency approach from Equation \eqref{eq:sal1} have been established, it is also possible to adopt improvements of the standard saliency explanations. For instance, Guided Backpropagation is a widely used explainability technique that uses gradient information \citep{guidebackprop}. Guided Backpropagation differs from Equation \eqref{eq:sal1} by zeroing out negative gradients in the backward pass of the backpropagation scheme. We define the Guided Backpropagation procedure for representations as:

\begin{equation}\label{eq:gb}
    \mathbf{S}_{{\textsc{gb}}} = \frac{1}{D}\sum\limits_{d=1}^D \nabla_{{\textsc{gb}}} f(\mathbf{X})_d.
\end{equation}
Second, SmoothGrad is another gradient-based explainability method that can be adopted from Equation \ref{eq:sal1} \citep{smoothGrad}. SmoothGrad injects noise into the input and produces an explanation by averaging over multiple explanations. We define SmoothGrad for representation as:

\begin{equation}\label{eq:sg}
    \mathbf{S}_{{\textsc{sg}}} = \frac{1}{M}\sum\limits_{m=1}^M\frac{1}{D}\sum\limits_{d=1}^D \nabla f(\mathbf{X}_m)_d,
\end{equation}
where $M$ is the number of explanations computed based on the noisy input.

\begin{figure*}[htb]
    \centering
    \includegraphics[width=\linewidth]{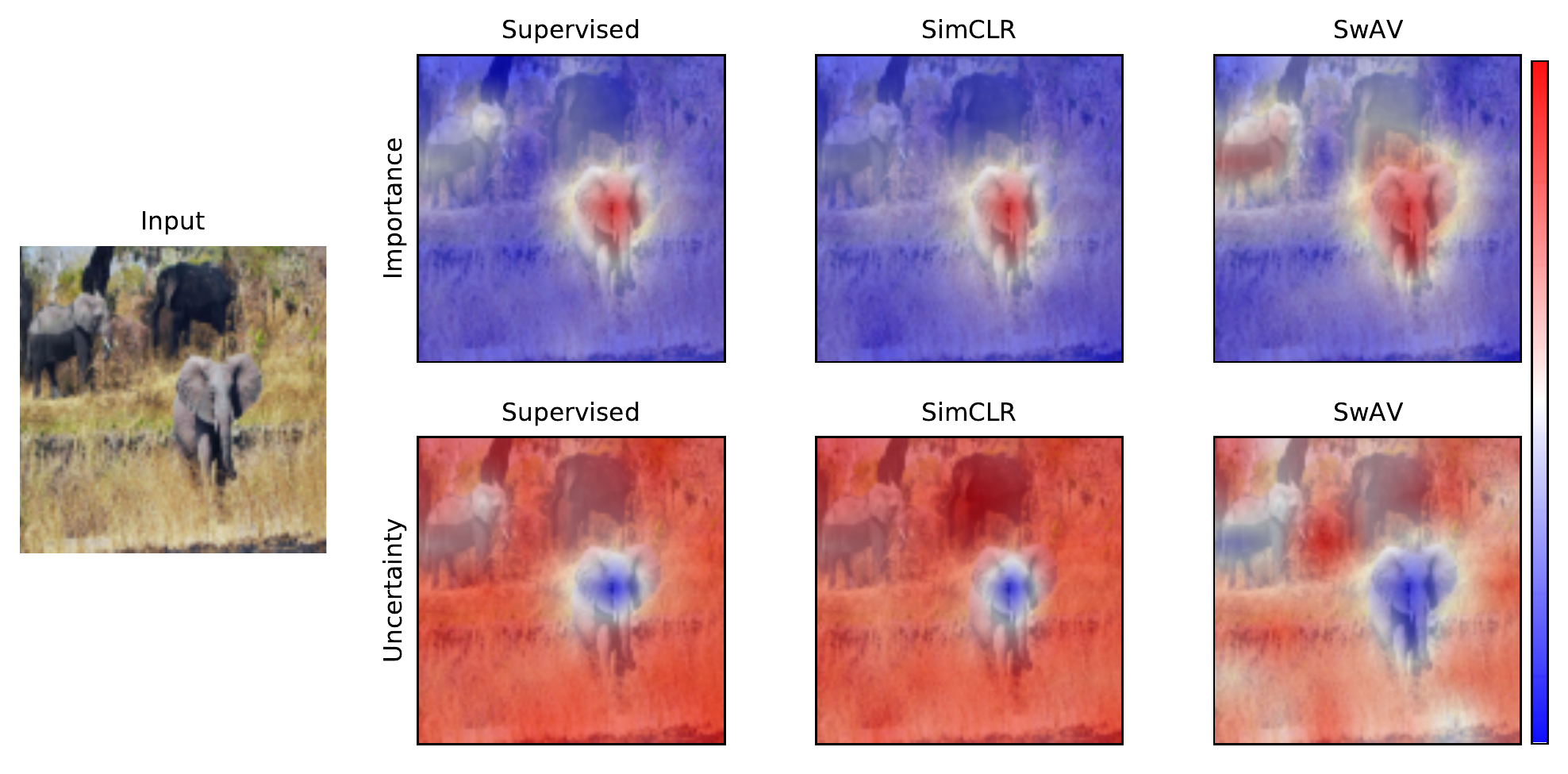}
    \caption{The figure shows the RELAX explanation and its uncertainty for the representation of the leftmost image for a number of widely used feature extractors. The first row displays the explanations for the representation and the second row shows the uncertainty associated with the different explanations. Red indicates high values and blue indicates low values. In this example, three elephants are visible in the image. The results show that all models highlight the elephant in the foreground as important for the representation, but there is more disagreement about the elephants in the background. Moreover, the uncertainty of the explanation for the elephant in the foreground is very low compared to the remaining regions of the image. Image is taken from MS COCO.}
    \label{fig:Ex1}
\end{figure*}

\section{Experiments}

To evaluate RELAX, we conduct numerous experiments and report both quantitative and qualitative results. We evaluate several features extraction models, both deep and non-deep, and trained with and without supervision. Our experiments show the advantageous of RELAX compared to the baselines, and illustrates how RELAX enables new approaches for analysing and understanding representation learning.

\noindent \textbf{Implementation details}. For the supervised model, we use the pretrained model from Pytorch \citep{pytorch}. For the models trained without labels but with self-supervision, we use the SimCLR \citep{simclr} and SwAV \citep{swav} frameworks, both of which have seen recent widespread use. These methods are chosen to represent two major types of self-supervised learning frameworks, namely contrastive instance learning (SimCLR) and clustering-based learning (SwAV). For SimCLR and SwAV, we use the pretrained models from Pytorch Lightning Bolts \citep{falcon2020framework}. We use a ResNet50 \citep{7780459} as the backbone for the feature extractors, and all models are trained on ImageNet \citep{deng2009imagenet}.

Similarly as in previous works \citep{Fong2019ICCV, Schulz2020Restricting}, we use the test split of the PASCAL VOC07 (VOC) \citep{Everingham2009} and the validation split of MSCOCO2014 (COCO) \citep{Lin2014} for evaluating the localisation scores, since they contain information about the location of the objects in the images. For the faithfulness score, we use the validation set of ImageNet \citep{deng2009imagenet}. For all datasets, we randomly sample 1000 images for evaluation and repeat all experiments 3 times. We generate 3000 masks to ensure a low estimator error. We set $h=w=7$ and resize all images to $H=W=224$, as suggested by \cite{Zhang2017}. For the monotonicity score, we use Alexnet \citep{alexnet} as the classifier, as suggested by \cite{7552539}. The threshold for U-RELAX is determined with median aggregation and $\gamma=1.0$, based on the empirical evaluation conduced in Section \ref{sec:hyp-urelax}.

\begin{table*}[ht]
    \centering
    \begin{tabular}{clcccccc}
    \toprule
    {Scores} & {Methods} & \multicolumn{2}{c}{Supervised} & \multicolumn{2}{c}{SimCLR}&  \multicolumn{2}{c}{SwAV}
    \\\cmidrule(lr){3-4}\cmidrule(lr){5-6}\cmidrule(lr){7-8}
    & {} &            VOC &            COCO &        VOC &        COCO &       VOC &      COCO \\
    \midrule
    \multirow{5}{*}{\centering \shortstack{pointing\\ game}}
    & Saliency               &          67.1±0.0 &         82.8±0.0 &    59.9±0.0 &   75.9±0.0 &      60.0±0.0 &     76.3±0.0 \\
    & Smooth Saliency        &          62.8±0.0 &         79.5±0.0 &    60.1±0.0 &   75.9±0.0 &      59.8±0.0 &     76.4±0.0 \\
    & Guided Saliency        &          66.6±0.0 &         82.9±0.0 &    58.4±0.0 &   73.3±0.0 &      59.5±0.0 &     75.8±0.0 \\
    & RELAX                  &          \textbf{72.6±0.1} &         \textbf{86.6±0.2} &    \textbf{68.7±0.3} &   \textbf{85.2±0.3} &      \textbf{67.8±0.2} &     \textbf{84.7±0.2} \\
    & U-RELAX  & 72.1±0.3 &     86.4±0.4&    68.6±0.2 &   85.0±0.5 &     66.7±0.7 &     84.1±0.4 \\
    \midrule
    \multirow{5}{*}{\centering top k}
    & Saliency               &          62.2±0.0 &         80.1±0.0 &    56.5±0.0 &   71.3±0.0 &      56.5±0.0 &     71.4±0.0 \\
    & Smooth Saliency        &          59.2±0.0 &         74.1±0.0 &    56.4±0.0 &   71.1±0.0 &      56.4±0.0 &     71.3±0.0 \\
    & Guided Saliency        &          62.2±0.0 &         80.2±0.0 &    55.1±0.0 &   69.0±0.0 &      56.3±0.0 &     71.1±0.0 \\
    & RELAX                  &          \textbf{72.8±0.4} &         \textbf{86.9±0.1} &    \textbf{69.0±0.3} &   \textbf{85.6±0.2} &      \textbf{68.1±0.4} &     \textbf{85.1±0.2} \\
    & U-RELAX  &    72.2±0.4 &         86.5±0.2 &    68.8±0.4 &   85.3±0.1 &      66.6±0.4 &     84.2±0.3 \\
    \midrule
    \multirow{5}{*}{\centering \shortstack{relevance\\ rank}}
    & Saliency               &          46.8±0.0 &         59.5±0.0 &    41.2±0.0 &   53.6±0.0 &      40.9±0.0 &     53.4±0.0 \\
    & Smooth Saliency        &          42.6±0.0 &         54.6±0.0 &    41.1±0.0 &   53.4±0.0 &      40.9±0.0 &     53.3±0.0 \\
    & Guided Saliency        &          46.8±0.0 &         59.8±0.0 &    40.6±0.0 &   53.0±0.0 &      40.9±0.0 &     53.3±0.0 \\
    & RELAX                  &          \textbf{56.4±0.0} &         \textbf{70.2±0.1} &    \textbf{54.2±0.2} &   \textbf{69.8±0.1} &      \textbf{52.4±0.1} &     \textbf{69.1±0.0} \\
    & U-RELAX  &          52.4±0.0 &         64.7±0.1 &    50.7±0.1 &   63.3±0.1 &      46.2±0.1 &     59.5±0.0 \\
    \bottomrule
    \end{tabular}
    \caption{Pointing game, top k, and relevance rank scores in percentages and averaged over 3 runs. Higher is better and bold numbers highlight the top performance. Results show that our method improves on the baseline across all scores.}
    \label{tab:localisation}
\end{table*}

\subsection{Qualitative Results}\label{sec:Qualresults}

Figure \ref{fig:FirstPage} and \ref{fig:Ex1} displays the explanation and the uncertainty in the explanations provided by RELAX for an image from the PASCAL VOC and MS COCO dataset, respectively. See Appendix \ref{morequalitatitive} for additional qualitative results. The input to the feature extractors is shown on the left, the first row shows the explanations, and the second row shows the uncertainties.

\subsubsection*{Are all instances of the same object equally important?}
Figure \ref{fig:FirstPage} shows an example with two objects, one bird prominently displayed in the foreground, and another more inconspicuous bird in the background. An interesting question that RELAX allows us to answer is: are both of these birds important for the representation of this image? And, are both of them equally important? First, all models indicate that the bird in the foreground is important, and that the explanations for this bird have low uncertainty. Second, SimCLR puts little emphasis on the bird in the background. In contrast, both the supervised feature extractor and SwAV are highlighting the second bird as having an influence on the representation. However, the uncertainty estimates for the second bird is slightly higher than those of the first bird, but still low compared to the remaining parts of the image.

\subsubsection*{What features are important in complex images with numerous objects?}
Figure \ref{fig:Ex1} shows an image with 3 elephants, one in the foreground and two in the background. Additionally, the background is more diverse and the objects have different lighting and perspective. Again, RELAX enables investigation of interesting aspects of the representations, such as: are the models capable of recognizing all elephants and utilizing the information? Does the models focus on background information instead of the objects? All models highlight the elephant in the foreground as important with high certainty. However, there is little emphasis on the shaded elephant, and the associated region of the image also has a high degree of uncertainty. Both the supervised model and SwAV put some importance on the third elephant with some degree of certainty, while SimCLR uses little or no information about the third elephant.

In both Figure \ref{fig:FirstPage} and \ref{fig:Ex1}, the SwAV feature extractor is focusing on several regions in the input, but with some regions of high uncertainty. While it is difficult to say exactly why, we hypothesize that it can be related to its self-supervised training procedure. SwAV relies on matching image views to a set of prototypes. Therefore, different parts of the input can be related to different prototypes, which we conjecture can lead to SwAV considering several regions of the input.

\begin{table*}[htb]
    \centering
    \begin{tabular}{clccc}
    \toprule
    {Scores} & {Methods} &  Supervised &      SimCLR &    SwAV \\
    \midrule
    \multirow{5}{*}{\centering monotonicity}
    & Saliency               &   12.8±0.2 &  14.8±0.5 &  14.6±0.3 \\
    & Smooth Saliency        &   15.4±0.1 &  14.3±0.3 &  14.0±0.3 \\
    & Guided Saliency        &   15.3±0.3 &  15.3±0.2 &  14.2±0.6 \\
    & RELAX                  &   18.3±0.5 &  20.2±0.4 &  \textbf{21.3±0.4} \\
    & U-RELAX  &   \textbf{23.6±0.4} &  \textbf{22.9±0.1} &  18.3±0.6 \\
    \bottomrule
    \end{tabular}
    \caption{Monotonicity scores averaged over 3 runs. Higher is better and bold numbers highlight the top performance. Results show that our method improves on the baseline.}
    \label{tab:faithfulness}
\end{table*}

\subsection{Quantitative Results}

Table \ref{tab:localisation} and  \ref{tab:faithfulness} displays the quantitative evaluation of our proposed methodology  compared with the gradient-based baselines described in Section \ref{sec:baseline}. The results show how the proposed method outperforms the baselines across all scores. The low standard deviation for RELAX show that the proposed methodology is robust to the stochasticity in the masks. Furthermore, the feature extractor trained using supervised learning achieves the highest performance compared to the feature extractors trained using self-supervised learning, which illustrates that label information does provide additional useful information for these scores.

For the localisation scores, RELAX provides the highest performance. The segmentation masks or boundix boxex can in many cases be large, and U-RELAX might remove uncertain points close to the boundaries of the segmentation masks. This might be desirable from a human perspective, as it provides clearer explanations with less uncertainty, but it will decrease the localisation scores. For the faithfulness score, U-RELAX provides a significant boost in performance for two encoders. The removal of uncertain explanations allows the classifier to focus on a smaller subset of highly relevant features. This can lead to the classifier having a more stable decrease in accuracy and a higher faithfulness score.

\subsection{Human Evaluation}

\begin{table*}[ht]
    \centering
    \begin{tabular}{lcccccc}\toprule
    {} & RELAX & U-RELAX & Saliency & Smooth Saliency & Guided Saliency & Random \\\midrule
    Counts & 79 & 29 & 9 & 4 & 8 & 1 \\\bottomrule
    \end{tabular}
    \caption{Human evaluation of representation explainability methods across 10 images from the PASCAL VOC dataset. Results show that the majority of the votes were cast for RELAX and U-RELAX.}
    \label{tab:human_eval}
\end{table*}

The localisation and faithfulness scores are both proxies for human evaluation that allow for quantitative analysis. However, the ultimate goal of XAI is to provide explanations that are understandable for people and align well with human intuition. Therefore, we conduct a user study with human evaluation of explanations. In this user study, 13 people were asked to select their preferred explanation from a selection of explanations across 10 different images. See Appendix \ref{humaneval} for a detailed description of the user study.

Table \ref{tab:human_eval} reports the results of the human evaluation. The results clearly indicate that RELAX and U-RELAX were the methods that aligned most closely with human intuition. Some participants highlighted that when both RELAX and the gradient-based methods indicated an object as important, they often preferred the more object focused explanation of RELAX, as opposed to the more edge focused explanations of the baselines. It was also noted that for some images the participants disagreed with most explanations, and would have provided a different explanation if possible. We believe that these are valuable insights that will be useful for improving explainability methods and also for designing future user studies.

\subsection{U-RELAX Hyperparameter Evaluation}\label{sec:hyp-urelax}

Table \ref{tab:loc-urelax} and \ref{tab:faith-urelax} reports localisation and faithfulness scores for different values of the hyperparameters in U-RELAX. Mean versus median aggregation is considered, and a selection of values for $\gamma$. The results indicate that setting $\gamma$ to less than 1, typical degrades performance. This can be understood by the thresholding being to strict and removing to many pixel indicated as important. Also, the differences between mean and median aggregation of the uncertainties is mostly low, but median aggregation gives a slight improvement, particularly for the relevance rank score and the monotonicity score.

\begin{table*}[ht]
    \centering
    \begin{tabular}{clcccccc}
    \toprule
    {Scores} & {(aggregation, $\gamma$)} & \multicolumn{2}{c}{Supervised} & \multicolumn{2}{c}{SimCLR}&  \multicolumn{2}{c}{SwAV}
    \\\cmidrule(lr){3-4}\cmidrule(lr){5-6}\cmidrule(lr){7-8}
    & {} &            VOC &            COCO &        VOC &        COCO &       VOC &      COCO \\
    \midrule
    \multirow{7}{*}{\centering \shortstack{pointing\\ game}}
    &  (mean, 0.95)   &          71.1±0.4 &         86.5±0.2 &    67.6±0.1 &   83.9±0.3 &      63.3±0.7 &     81.1±0.5 \\
    &  (mean, 0.99)   &          71.8±0.4 &         86.4±0.5 &    68.6±0.4 &   \textbf{85.0±0.4} &      66.4±0.6 &     \textbf{84.2±0.4} \\
    &  (mean, 1.0)    &          71.7±0.1 &         86.5±0.2 &    68.6±0.1 &   \textbf{85.0±0.3} &      \textbf{66.7±0.7} &     84.1±0.2 \\
    &  (median, 0.95) &          71.2±0.2 &         \textbf{86.6±0.1} &    67.6±0.4 &   84.2±0.2 &      63.6±0.2 &     80.9±0.1 \\
    &  (median, 0.99) &          71.8±0.3 &         86.5±0.4 &    \textbf{68.8±0.3} &   \textbf{85.0±0.2} &      66.3±0.6 &     84.0±0.3 \\
    &  (median, 1.0)  &          \textbf{72.1±0.3} &         86.4±0.4 &    68.6±0.2 &   \textbf{85.0±0.5} &      \textbf{66.7±0.7} &     84.1±0.4 \\
    \midrule
    \multirow{7}{*}{\centering \shortstack{top k}}
    &  (mean, 0.95)   &          71.3±0.4 &         86.2±0.2 &    67.1±0.1 &   83.2±0.3 &      62.8±0.2 &     79.5±0.4 \\
    &  (mean, 0.99)   &          \textbf{72.2±0.4} &         \textbf{86.6±0.2} &    \textbf{68.8±0.3} &   85.2±0.2 &      66.4±0.2 &     84.0±0.3 \\
    &  (mean, 1.0)    &          \textbf{72.2±0.4} &         86.5±0.2 &    \textbf{68.8±0.4} &   \textbf{85.3±0.1} &      \textbf{66.7±0.4} &     \textbf{84.3±0.2} \\
    &  (median, 0.95) &          71.2±0.4 &         86.1±0.2 &    67.1±0.2 &   83.2±0.4 &      62.7±0.2 &     79.1±0.4 \\
    &  (median, 0.99) &          \textbf{72.2±0.4} &         86.5±0.2 &    68.7±0.3 &   85.2±0.2 &      66.4±0.2 &     83.9±0.3 \\
    &  (median, 1.0)  &          \textbf{72.2±0.4} &         86.5±0.2 &    \textbf{68.8±0.4} &   \textbf{85.3±0.1} &      66.6±0.4 &     84.2±0.3 \\
    \midrule
    \multirow{7}{*}{\centering \shortstack{relevance\\ rank}}
    &  (mean, 0.95)   &          45.9±0.0 &         55.7±0.0 &    41.6±0.1 &   52.3±0.1 &      39.6±0.1 &     51.0±0.0 \\
    &  (mean, 0.99)   &          50.3±0.0 &         61.2±0.1 &    48.6±0.1 &   59.8±0.1 &      44.0±0.1 &     56.0±0.1 \\
    &  (mean, 1.0)    &          51.4±0.1 &         63.0±0.1 &    50.3±0.1 &   62.2±0.1 &      45.6±0.1 &     58.2±0.1 \\
    &  (median, 0.95) &          46.8±0.0 &         57.2±0.1 &    42.4±0.1 &   53.3±0.1 &      40.4±0.1 &     52.1±0.0 \\
    &  (median, 0.99) &          51.2±0.0 &         63.0±0.1 &    49.1±0.1 &   60.8±0.1 &      44.6±0.1 &     57.3±0.1 \\
    &  (median, 1.0)  &          \textbf{52.4±0.0} &         \textbf{64.7±0.1} &    \textbf{50.7±0.1} &   \textbf{63.3±0.1} &      \textbf{46.2±0.1} &     \textbf{59.5±0.0} \\
    \bottomrule
    \end{tabular}
    \caption{Evaluation of U-RELAX hyperparameters in terms of pointing game, top k, and relevance rank scores in percentages and averaged over 3 runs. Higher is better and bold numbers highlight the top performance}
    \label{tab:loc-urelax}
\end{table*}

\begin{table*}[htb]
    \centering
    \begin{tabular}{clccc}
    \toprule
    {Scores} & {(aggregation, $\gamma$)} &  Supervised &  SimCLR &    SwAV \\
    \midrule
    \multirow{6}{*}{\centering monotonicity}
    & (mean, 0.95)   &   16.3±0.5 &  11.8±0.3 &  12.4±0.3 \\
    & (mean, 0.99)   &   22.2±0.2 &  20.4±0.5 &  16.2±0.3 \\
    & (mean, 1.0)    &   23.2±0.1 &  21.8±0.3 &  18.0±0.0 \\
    & (median, 0.95) &   17.9±0.7 &  12.8±0.2 &  13.5±0.2 \\
    & (median, 0.99) &   23.0±0.7 &  21.1±0.1 &  17.1±0.4 \\
    & (median, 1.0)  &   \textbf{23.6±0.4} &  \textbf{22.9±0.1} &  \textbf{18.3±0.6} \\
    \bottomrule
    \end{tabular}
    \caption{Evaluation of U-RELAX hyperparameters in terms of monotonicity score in percentages and averaged over 3 runs. Higher is better and bold numbers highlight the top performance}
    \label{tab:faith-urelax}
\end{table*}

\subsection{Use Case I: Multi-View Clustering}
\begin{figure}
    \centering
    \includegraphics[width=\linewidth]{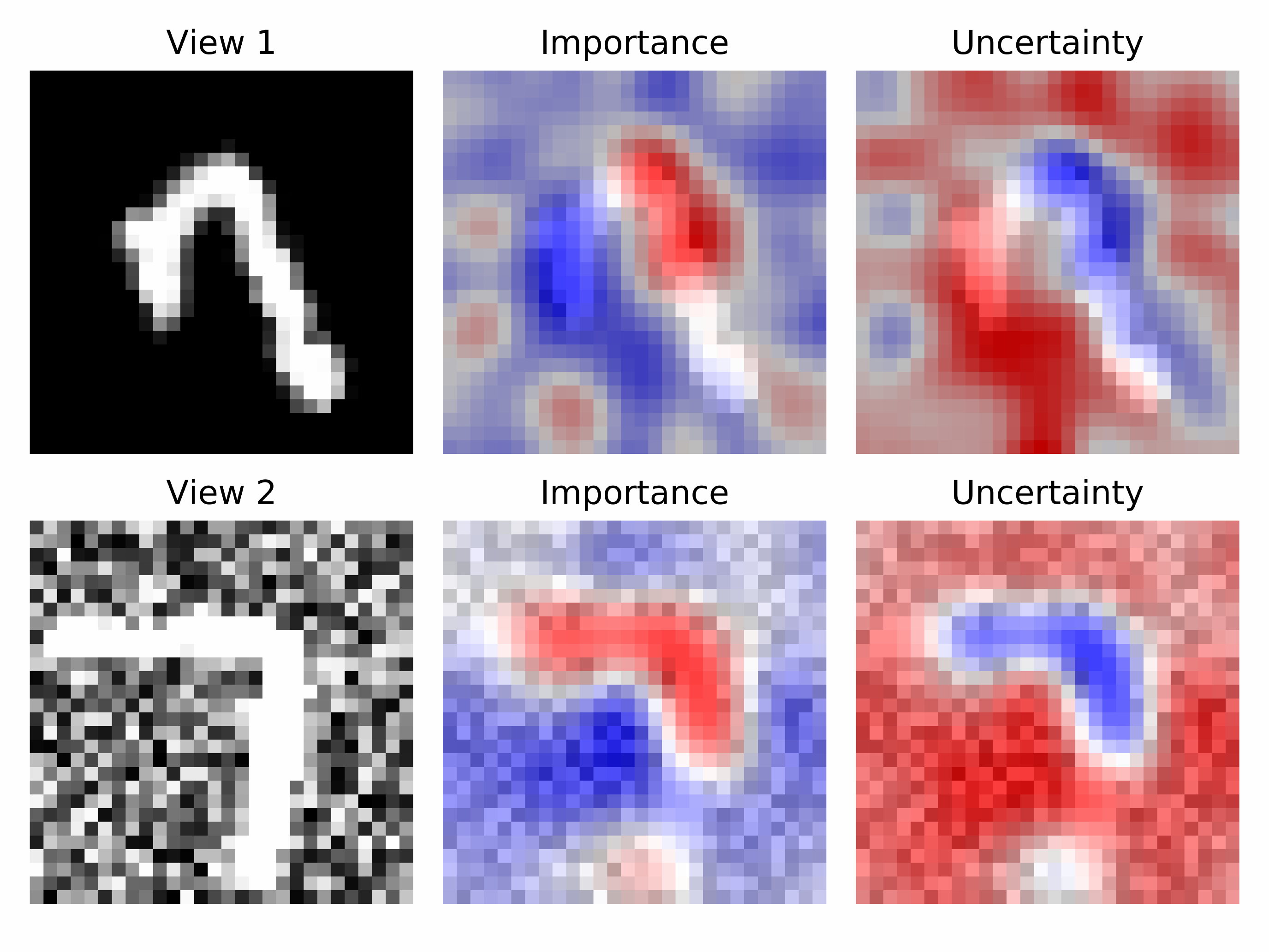}
    \caption{RELAX explanation and uncertainty for the representation of an example from Noisy MNIST image for a number of widely used feature extractors. The first row displays input, explanation, and uncertainty for view 1, and the second row for view 2. Red indicates high values and blue indicates low values. The Figure shows that Completer is extracting complementary information from the two views for creating its unified representation. \vspace{-0.2cm}}
    \label{fig:completer}
\end{figure}

To further illustrate the ability of RELAX to obtain insights into new tasks, we conduct an experiment on multi-view clustering. We learn a feature extractor using the Completer framework \citep{completer}, which uses an information theoretic approach to fuse several views into a new representation. Completer uses individual encoders for each view, and concatenates the representation from each encoder to produce a unified representation. Clustering is performed by applying K-means to the learned representations. To adopt RELAX for such a setting, we generate individual masks for each view and monitor the change in the representation in the unified representation space. While there is no way to investigate which parts of the different views that influence the unified representation in the Completer framework, using RELAX allows us to answer this question. Figure \ref{fig:completer} shows an example on Noisy MNIST \citep{noisyMNIST}, where one view is a digit and the other view is a noisy version of the same digit. The result shows that the Completer framework is exploiting information from both views to produce a new representation, even if one view contains more noise. Such insights would not be obtainable without RELAX.

\begin{figure*}[htb]
    \centering
    \includegraphics[width=0.975\linewidth]{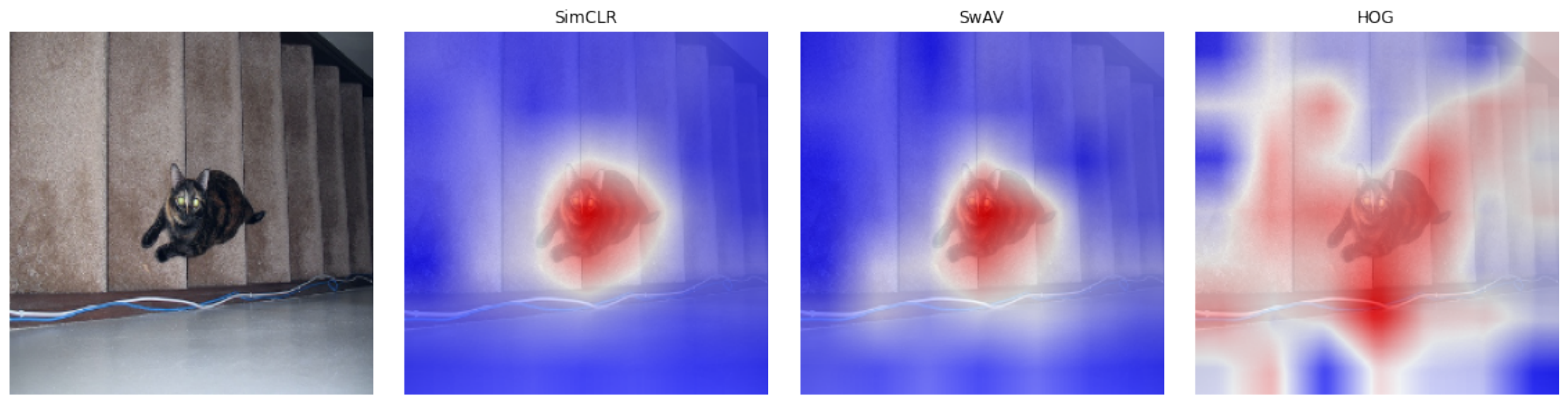}
    \caption{The figure shows the RELAX explanation for two deep learning-based feature extractors compared with the traditional HOG algorithm. Figure shows how HOG features focus on more indistinct regions in the input, while deep learning methods focus mainly on the cat. Image is taken from PASCAL VOC.}
    \label{fig:hog1}
\end{figure*}

\begin{figure*}[htb]
    \centering
    \includegraphics[width=0.975\linewidth]{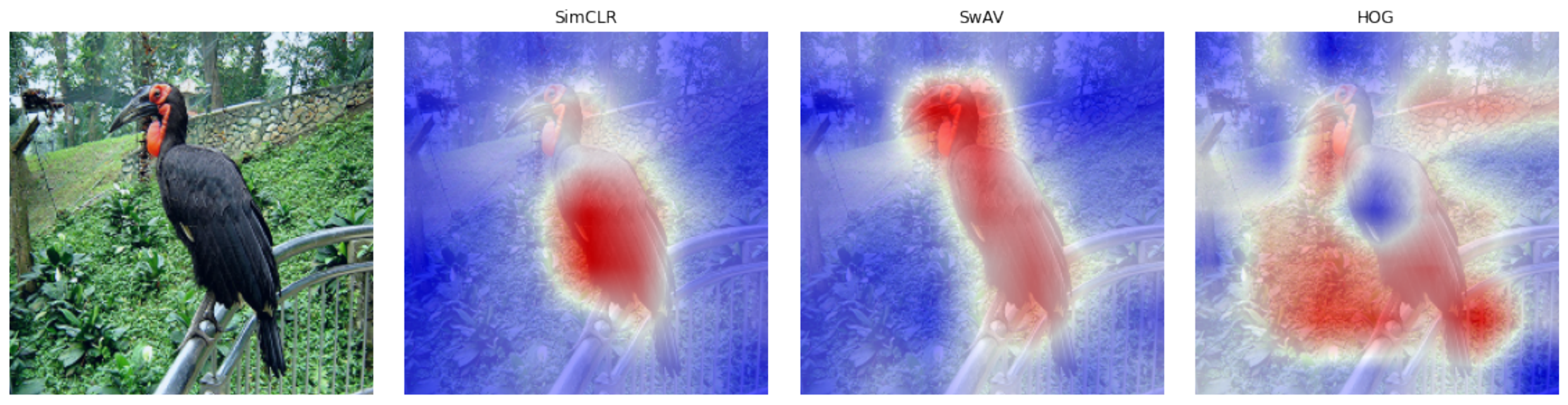}
    \caption{The figure shows the RELAX explanation for two deep learning-based feature extractors compared with the traditional HOG algorithm. Figure shows how HOG features puts little attention on the bird and mostly focus on the background. Image is taken from PASCAL VOC.}
    \label{fig:hog2}
\end{figure*}

\subsection{Use Case II: Explaining HOG Features}

RELAX is not limited to representations produced by deep neural networks. It can be used to explain the representation produced by any function that transform an image into a vector representation. To illustrate the versatility of RELAX, we explain representation produced by the Histogram of Oriented Gradients (HOG) feature extraction method \citep{1467360}, which have been used extensively in the computer vision literature. Figure \ref{fig:hog1} and \ref{fig:hog2} shows two examples where the explanation for the HOG representation is compared with the SimCLR and SwAV representations. We consider the representations from these two methods since they are also unsupervised like the HOG features. 

Features produced by deep neural networks are typically allow for higher performance than those from algorithms such as HOG and other handcrafted feature extraction methods. RELAX provides insights into why this is. In Figure \ref{fig:hog1}, both the SimCLR and the SwAV feature extractors focus on the cat in the center of the images. The HOG algorithm has a more widespread focus. Also, much of the emphasis is put on the cord going along the staircase. Since the HOG algorithm is utilizing gradient information, these sharp lines will have a big influence on the representation, and it is therefore not surprising that the cat receives less attention. In Figure \ref{fig:hog1}, both SimCLR and SwAV focus on the bird, while the HOG features are more focused on other regions in the image. For instance, the iron rod and a tree in the background and are indicated as being important for the representation of this image. Both examples provide insights into why HOG features lead to inferior performance, when compared with features produced by deep neural networks. This information would not be available without the proposed RELAX framework.

\section{Conclusion}
In this work, we presented RELAX, a framework for explaining representations produced by any feature extractors. RELAX is based on masking out parts of an image and measuring the similarity with an unmasked version in the representation space. We introduced a principled approach to quantifying uncertainty in explanations. RELAX was evaluated by comparing several widely used feature extractors. Results indicate that there can be a big difference in the quality of the explanations. It was shown that filtering out parts of an explanation based on its uncertainty can improve the faithfulness, and that RELAX can have a facilitating role, providing explainability for several downstream applications such as multi-view clustering. We believe that RELAX can be an important addition in the intersection between XAI and representation learning.

\backmatter

\bmhead{Acknowledgments}

The authors would like to thank Nils Midtbø, Caroline Granås, Theodor Ross, Suaiba Salahuddin, Sigrid Vold Jensen, Thomas Johansen, Julianne Nyvold, Kristoffer Furøy, Jostein Henriksen, Erland Grimstad, Jonas Mørch-Lampe, Inger Solheim, and Andreas Kvammen for participating in the user study through human evaluation.

\section*{Statements and Declarations}

\subsection{Competing Interests}

The authors have no competing interests to declare that are relevant to the content of this article.

\subsection{Funding}

This work was financially supported by the Research Council of Norway (RCN), through its Centre for Research-based Innovation funding scheme (Visual Intelligence, grant no. 309439), and Consortium Partners. The work was further partially funded by RCN FRIPRO grant no. 315029, RCN IKTPLUSS grant no. 303514, and the UiT Thematic Initiative “Data-Driven Health Technology”.

\begin{appendices}

\section{}\label{distshift}

An alternative approach for creating the random variable $\bar{\mathbf{h}}$ is the following:

\begin{equation}
    \bar{\mathbf{h}} = f(\mathbf{X} \odot \mathbf{M}-\mathbf{D}(1-\mathbf{M})),
\end{equation}

where each element of $\mathbf{D}$ follows $N(\mu_{x_{ij}}, \sigma_{x_{ij}})$. The mean $\mu_{x_{ij}}$ and standard deviation $\sigma_{x_{ij}}$ is estimated by averaging across all samples in the data. Such a strategy could avoid potential distribution shifts that might occur when zeroing out large parts of the image, but also required determining the mean and variance of the data distribution.

Table \ref{tab:loc-dist} displays localisation scores scores the two masking strategies outlines in Section \ref{sec:relax}, namely zero masking or insertion of normally distributed noise. While there is some variation in the results, masking out with zeros provide the highest performance overall.

\begin{table*}[ht]
    \centering
    \begin{tabular}{clcccccc}
    \toprule
    {Scores} & {Methods} & \multicolumn{2}{c}{Supervised} & \multicolumn{2}{c}{SimCLR}&  \multicolumn{2}{c}{SwAV} \\
    \midrule
    & {} &            VOC &            COCO &        VOC &        COCO &       VOC &      COCO
    \\\cmidrule(lr){3-4}\cmidrule(lr){5-6}\cmidrule(lr){7-8}
    \vspace{0.075cm}
    \multirow{2}{*}{\centering \shortstack{pointing\\ game}}
    & RELAX (zeros)                  &          \textbf{72.6±0.1} &         \textbf{86.6±0.2} &    \textbf{68.7±0.3} &   \textbf{85.2±0.3} &      \textbf{67.8±0.2} &     84.7±0.2 \\
    & RELAX (noise)          &          72.0±0.5 &         86.0±0.3 &    66.6±0.1 &   84.3±0.7 &      67.7±0.5 &     \textbf{85.1±0.3} \\
    \midrule
    \vspace{0.075cm}
    \multirow{2}{*}{\centering \shortstack{top k}}
    & RELAX (zeros)                  &          \textbf{72.8±0.4} &         \textbf{86.9±0.1} &    \textbf{69.0±0.3} &   \textbf{85.6±0.2} &      68.1±0.4 &     85.1±0.2 \\
    & RELAX (noise)          &          72.4±0.4 &         86.5±0.1 &    66.0±0.3 &   84.2±0.2 &      \textbf{68.2±0.3} &     \textbf{85.3±0.2} \\
    \midrule
    \vspace{0.1cm}
    \multirow{2}{*}{\centering \shortstack{relevance\\ rank}}
    & RELAX (zeros)                  &          56.4±0.0 &         \textbf{70.2±0.1} &    \textbf{54.2±0.2} &   \textbf{69.8±0.1} &      52.4±0.1 &     69.1±0.0 \\
    & RELAX (noise)          &          \textbf{56.7±0.0} &         70.1±0.1 &    53.5±0.1 &   68.5±0.0 &      \textbf{52.8±0.1} &     \textbf{69.2±0.0} \\
    \bottomrule
    \end{tabular}
    \caption{Evaluation of zero versus noise masking strategy in terms of pointing game, top k, and relevance rank scores in percentages and averaged over 3 runs. Higher is better and bold numbers highlight the top performance. Results indicate that zero masking provides the best performance.}
    \label{tab:loc-dist}
\end{table*}

\section{}\label{maskinstrat}

\begin{figure*}[htb]
    \centering
    \includegraphics[width=0.95\linewidth]{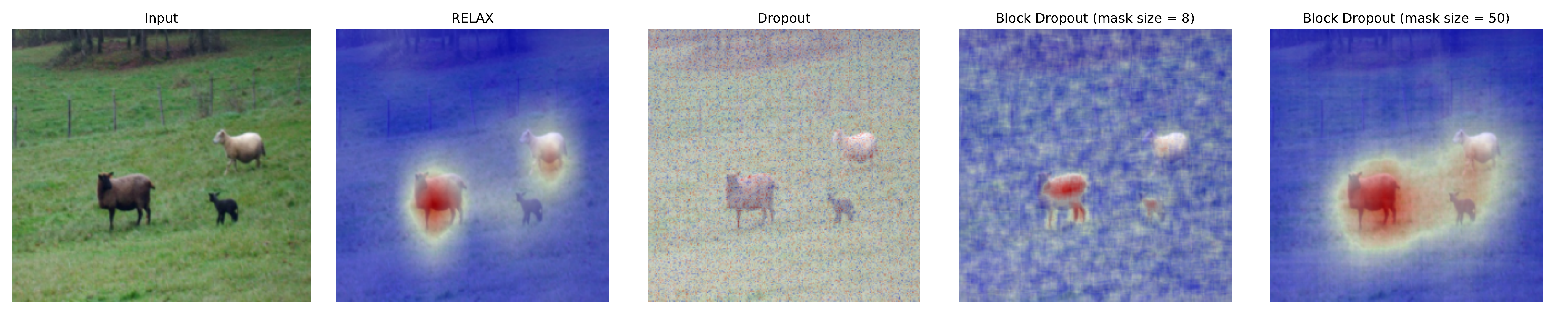}
    \caption{Comparison of different masking strategies. Leftmost image shows input, and second to left is the RELAX explanations with the masking presented in the main paper. The center image is with Bernoulli-noise (Dropout) directly on the input, and the remaining two images are with Block Dropout with different block size. The example illustrates that other masking strategies either fail completely, or require per-image parameter tuning, which is impractical in most scenarios.}
    \label{fig:mask_strat}
\end{figure*}

Figure \ref{fig:mask_strat} shows alternative strategies for masking out part of the input. One alternative is to apply Bernoulli noise to the input, which is equivalent to using Dropout \citep{jmlrdrop} on the input. However, However, this does not introduce noise with spatial awareness, and therefore results in failing to explain the representation of the image.
Another option is to drop regions of the input, such that objects could be fully or partially removed from the input. This could be achieved using the DropBlock algorithm \citep{dropBlock}. However, this requires tuning the size of the mask on the input, which will be highly dependent on the objects present in the image. Such a per-image tuning would be impractical in most scenarios.

\section{}\label{proofs}

In this section we present the proofs for all theorems in the main paper.

\subsection{Proof of Theorem \ref{teo:masks}}
\begin{proof}
    First, let the Bounded difference assumption be defined as follows:
    \begin{definition}[Bounded difference assumption]
        Let a be some set and $f: A^N$ $f:A^N \to \mathbb{R}$. The function $f$ satisfies the bounded differences assumption if if there exists real numbers $c_1,\ldots,c_N \geq 0$ so that for all $i=1,\ldots,N$,
        \begin{equation}
            \sup\limits_{x_1,\ldots,x_N, x_i \in A} \lvert f(x_1,\ldots,x_N, x'_i)- f(x_1,\ldots,x_N, x'_i)\rvert
        \end{equation}
    \end{definition}
    
    We then have the following lemmas:
    \begin{lemma}[McDiarmid's inequality]\label{lem:mcdiarmid}
        Let $X_1,\ldots,X_N$ be arbitrary independent random variables on set A and $f:A^N \to \mathbb{R}$ satisfies the bounded difference assumption. Then, for all $t>0$
        \begin{equation}
            \begin{split}
                P(\lvert f(X_1,\ldots,X_N)-\mathrm{E}[f(X_1,\ldots,X_N)\rvert] & \geq t) \\ \leq 
                & 2e^{\frac{-2t^2}{\sum_{n=1}^N c_n^2}}
            \end{split}
        \end{equation}
    \end{lemma}
    \begin{proof}
        See \cite{mcdiarmid1989}.
    \end{proof}
    \begin{lemma}\label{lem:mcdiarmid2}
        Let $X_1,\ldots,X_N$ and $f$ be defined as in Lemma \ref{lem:mcdiarmid}, then if each $X_n$ satisfies $X_n \in (a_n, b_n)$ and $f(X_1,\ldots,X_N)=\sum_{n=1}^N X_n$, then $c_n=b_n-a_n$.
    \end{lemma}
    \begin{proof}
        See \cite{mcdiarmid1989}.
    \end{proof}
    
    We are now ready to prove the theorem. First, let
    \begin{align}
        X_n=\frac{s(\mathbf{h}, \bar{\mathbf{h}}_n) M_{ij}(n)}{N},
    \end{align}
    and 
    \begin{align}
        f(X_1,\ldots,X_n) = \sum_{n=1}^{N}X_n.
    \end{align}
    Since $s(\cdot, \cdot)$ is bounded in $(0, 1)$ (we use the cosine similarity between vectors with non-negative elements (ReLU outputs)), we have $a_n = 0$ and $b_n = 1/N$, which gives $c_n = 1/N$ by Lemma \ref{lem:mcdiarmid2}.
    
    Now, observe that
    \begin{align}
        f(X_1,\ldots,X_n) = \frac{1}{N}\sum\limits_{n=1}^{N} s(\mathbf h, \bar{\mathbf h}_n) M_{ij}(n) = \bar R_{ij}.
    \end{align}
    
    Combining Lemmas \ref{lem:mcdiarmid} and \ref{lem:mcdiarmid2} then gives
    \begin{align}
        P(\lvert \bar R_{ij}- R_{ij} \rvert] & \geq t) \leq  2e^{\frac{-2t^2}{\sum_{n=1}^N (1/N)^2}}
    \end{align}
    for all $t > 0$. Inserting $N = - \ln(\delta / 2)/2t^2$ gives
    \begin{align}
        P(\lvert \bar R_{ij}- R_{ij} \rvert] \geq t) & \leq  2e^{\frac{-2t^2}{\sum_{n=1}^N (1/N)^2}} \\
        & = 2 e^{-2t^2 \left(- \frac{\ln(\delta/2)}{2 t^2} \right) } \\
        & = 2 e^{\ln(\delta / 2)} \\
        & = \delta,
    \end{align}
    which concludes our proof.
\end{proof}

\begin{figure}[htb]
    \centering
    \includegraphics[width=\linewidth]{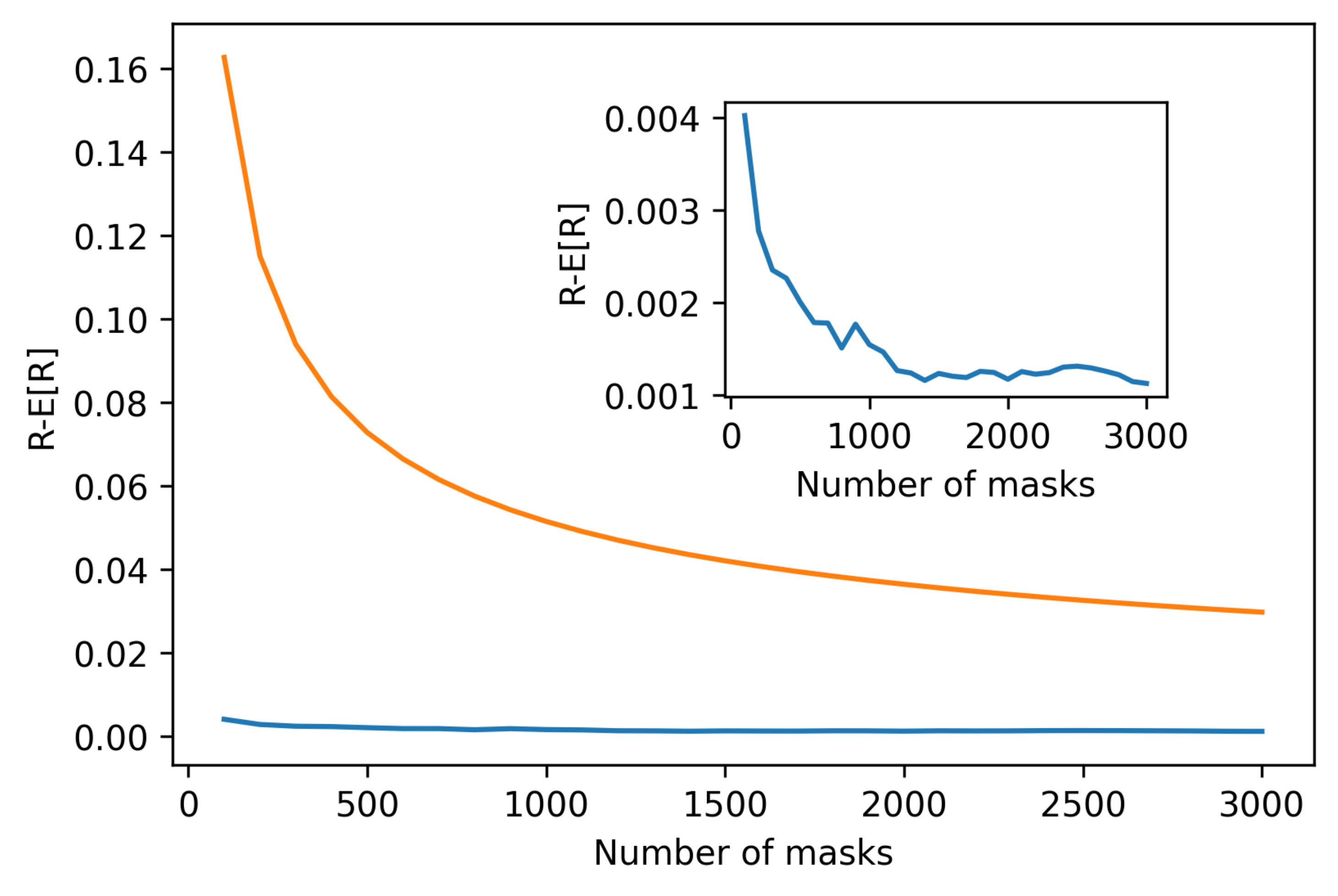}
    \caption{Empirical evaluation of the derived bound for the number of masks necessary for low estimation error. We calculate the absolute error as the number of masks increase, average over 10 randomly samples images from the PASCAL VOC dataset. To obtain a value for $R_{ij}$, we use 10000 masks and average over 10 runs for a single sample. Results indicate that the estimation error is much lower than the predicted bound.}
    \label{fig:Rbound}
\end{figure}

In Figure \ref{fig:Rbound} we show an empirical validation the bound. We calculate the absolute error as the number of masks increase, averaged over 10 randomly sampled images from the PASCAL VOC dataset. To obtain a value for $R_{ij}$, we use 10000 masks and average over 10 runs for a single sample. The results indicate that the true error is much lower than the proposed bound, which we attribute to setting $a_n=0$. While it is possible to obtain a similarity of 0, it is highly unlikely since our masking strategy never removes all information in an image.

\subsection{Proof of Theorem \ref{thm:rkhs}}

\begin{proof}
    Since $s(\cdot, \cdot)$ is a valid Mercer kernel, we can write $s(\mathbf h, \bar{\mathbf h}_n) = \langle \phi(\mathbf h), \phi(\bar{\mathbf h}_n) \rangle_\mathcal{H}$. This gives 
    \begin{align}
    \bar{R}_{ij} &= \frac{1}{N}\sum_{n=1}^N \langle \phi(\mathbf{h}), \phi(\bar{\mathbf{h}}_n)\rangle_{\mathcal H} M_{ij}(n) \\
     &= \langle \phi(\mathbf{h}), \frac{1}{N}\sum_{n=1}^N\phi(\bar{\mathbf{h}}_n) M_{ij}(n)\rangle_{\mathcal H}\label{eq:kern1}
\end{align}
by the bilinearity of the inner product on $\mathcal H$.
\end{proof}

\subsection{Proof of Theorem \ref{thm:parzen}}
\begin{proof}
    Observe that
    \begin{align}
        & \bar R_{ij} \cdot \frac{N}{\sum_{n'=1}^{N}M_{ij}(n')}\\ 
        & = \frac{N}{\sum_{n'=1}^{N}M_{ij}(n')} \cdot \frac{1}{N} \sum\limits_{n=1}^{N} s(\cdot, \bar{\mathbf h}_n) M_{ij}(n) \\
        & = \frac{1}{\sum_{n'=1}^{N}M_{ij}(n')} \sum\limits_{n=1}^{N} s(\cdot, \bar{\mathbf h}_n) M_{ij}(n) \\
        & = p_{ij}(\mathbf h)
    \end{align}
    $\bar R_{ij}$ is therefore proportional to $p_{ij}(\mathbf h)$.
\end{proof}

\section{}\label{supp:onevstwo}

\begin{figure}[htb]
    \centering
    \includegraphics[width=\linewidth]{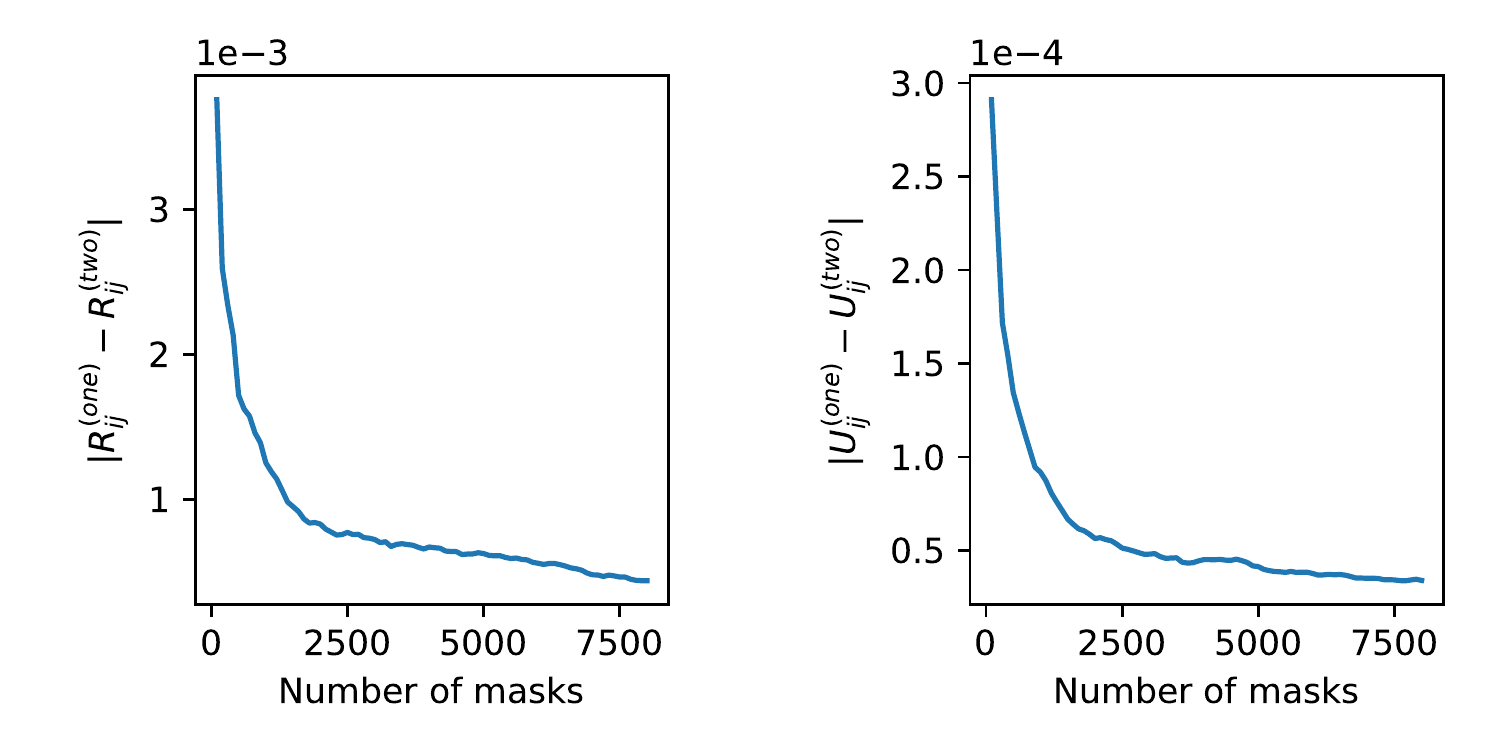}
    \caption{Absolute error of one-pass versus two-pass version of RELAX for importance (leftmost figure) and uncertainty (rightmost figure), averaged over 50 images from the VOC dataset. The figure shows how the difference between the versions is small for both the importance and uncertainty estimates.}
    \label{fig:1vs2}
\end{figure}

We investigate the potential differences between the one-pass and two-pass version of RELAX. For a given image, we calculate the absolute error between the one-pass and two-pass estimates for different number of masks. The results are shown in Figure \ref{fig:1vs2} and illustrate that the difference between the two methods is very small, particularly as the number of masks increases. However, since the one-pass version computes both the importance and uncertainty in one pass through the data, it requires only half the number of masks compared to the two pass version, thus increasing the computational efficiency of RELAX.

\section{}\label{humaneval}
\begin{figure*}[htb]
    \centering
    \includegraphics[width=0.75\linewidth]{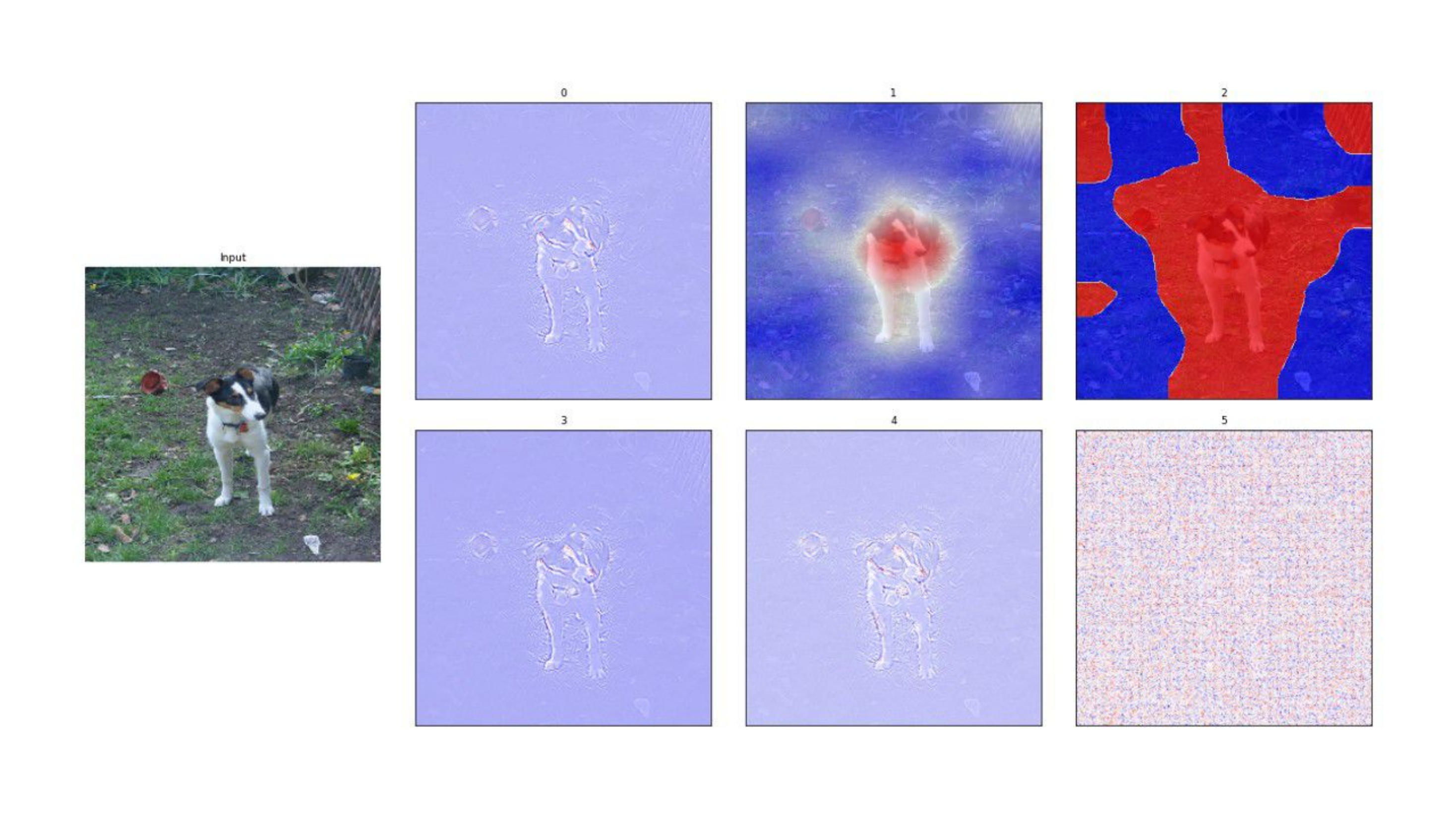}
    \caption{Example from the human evaluation experiment. Participants were asked to select which explanation they preferred out of the 6 alternatives. For each of the images, the explanations were shuffled in a random order. One of the explanations for each image was randomly sampled from random noise, in order to assess if any participants would select a nonsensical explanation.}
    \label{fig:xai_eval}
\end{figure*}
The user study in the main manuscript was conducted by having a group of participants select among competing explanations for a random selection of images from the PASCAL VOC dataset. The group of participants consisted of men and women, where some had knowledge of machine learning and other were uneducated. None of the participants have been involved in the development of this work. Figure \ref{fig:xai_eval} displays an example from the study. The participants were shown an image with 6 competing explanations, and asked to chose which one they preferred. To determine which explanation each participant judged to be the "best", they were told to ask themselves the following questions:

\begin{displayquote}
"Which of these explanations agree the most with how you would explain the important content in the given image?
\end{displayquote}
For each image, the explanations were shuffled randomly. The participants were shown 10 images, and asked to only pick on explanation. Overall, 13 people participated in the study.

There are several limitations. Both the number of images and the number of participants could have been greater. The participants had to chose one explanation, when in some cases they might have wanted to select none or more explanations. Also, the images could have been selected from other datasets. There are also potential biases with the study. Most participants are from one country and from a limited age segment. Lastly, we did not control the type of screen that participants performed their evaluation on, which could also have an undesirable affect. 

\section{}\label{morequalitatitive}

\begin{figure*}[htb]
    \centering
    \includegraphics[width=0.75\linewidth]{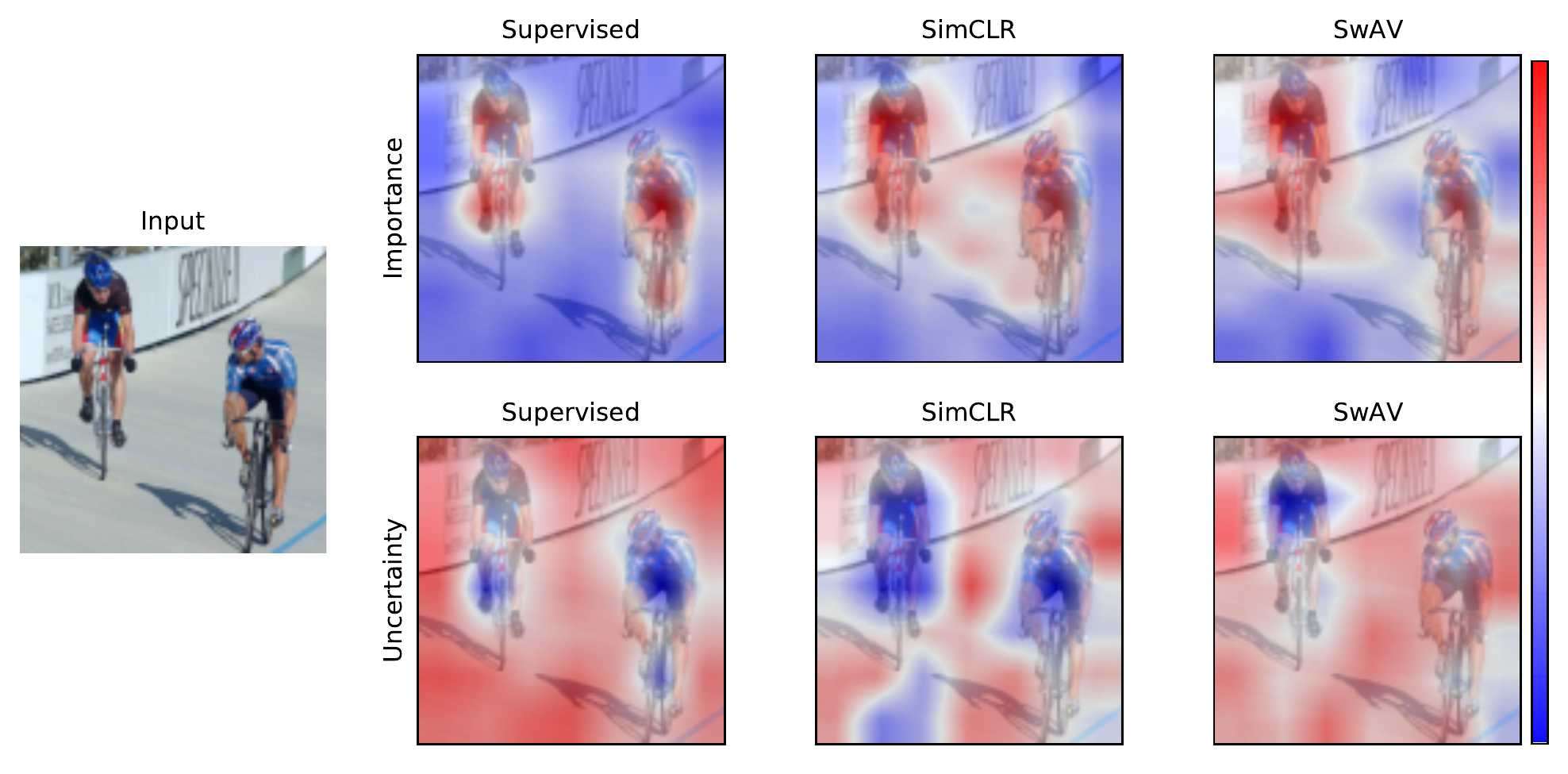}
    \caption{Example from the VOC dataset.}
    \label{fig:ex1}
\end{figure*}

\begin{figure*}[htb]
    \centering
    \includegraphics[width=0.75\linewidth]{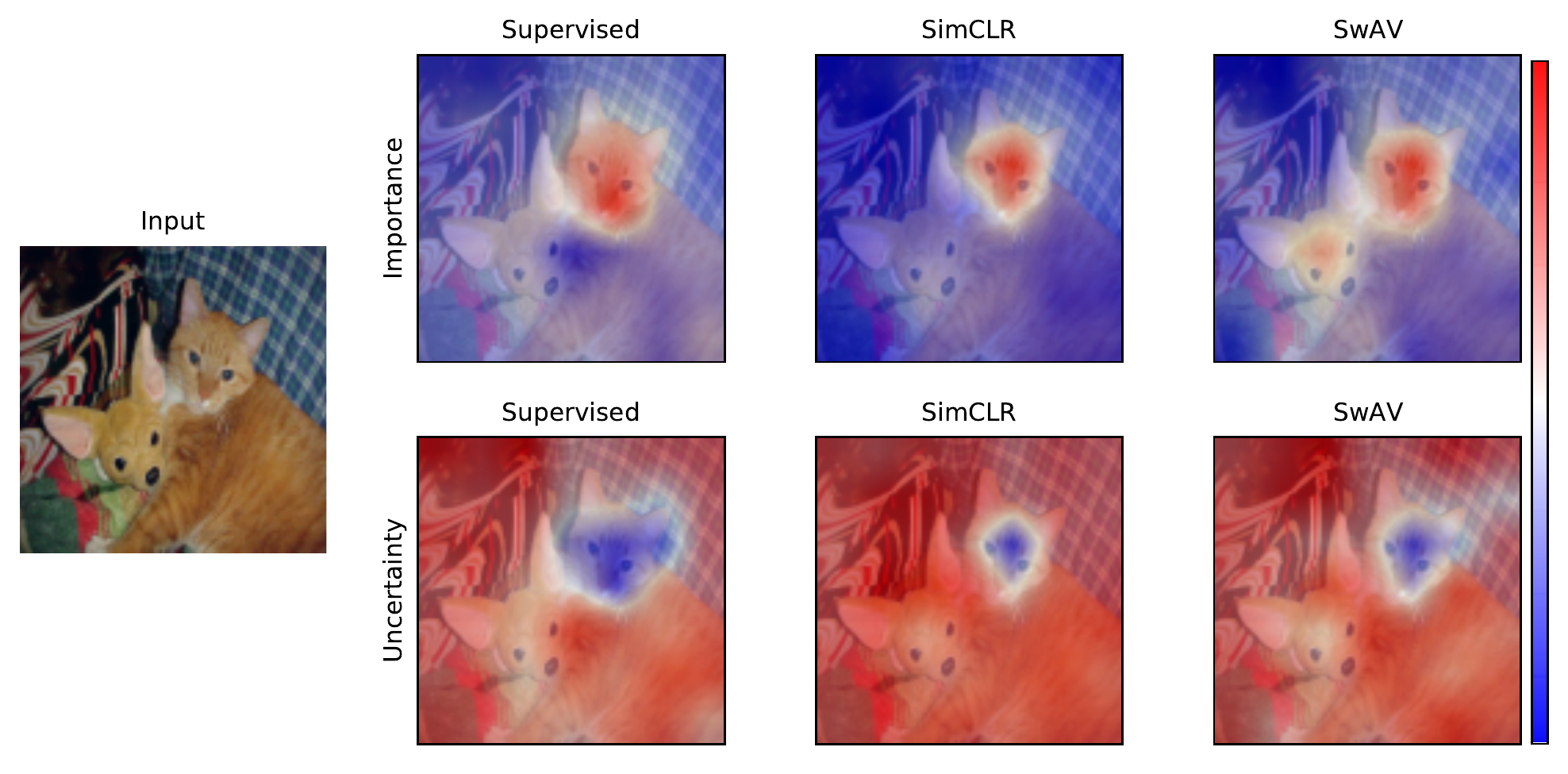}
    \caption{Example from the COCO dataset.}
    \label{fig:ex2}
\end{figure*}

\begin{figure*}[htb]
    \centering
    \includegraphics[width=0.75\linewidth]{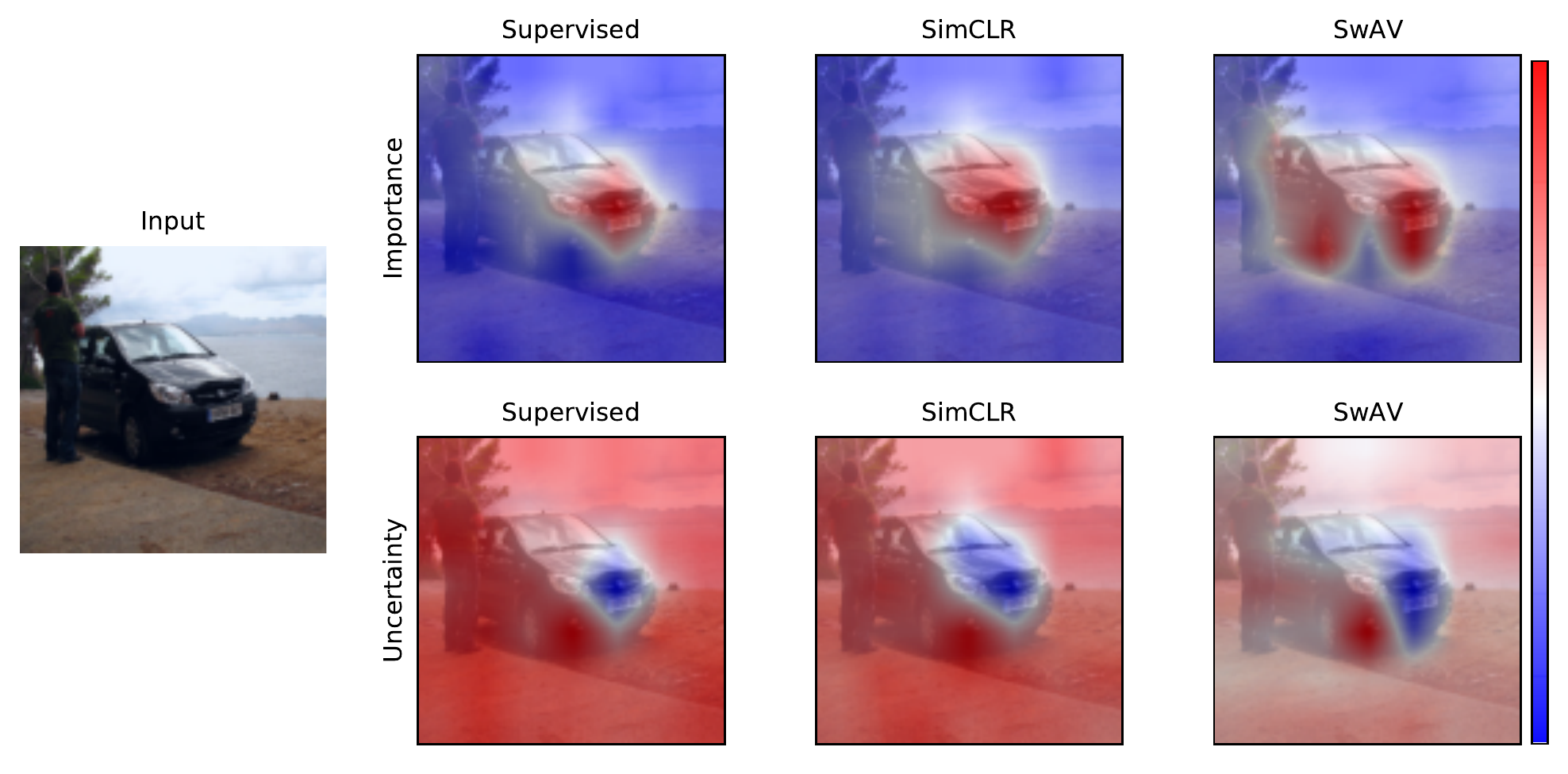}
    \caption{Example from the VOC dataset.}
    \label{fig:ex3}
\end{figure*}

\begin{figure*}[htb]
    \centering
    \includegraphics[width=0.75\linewidth]{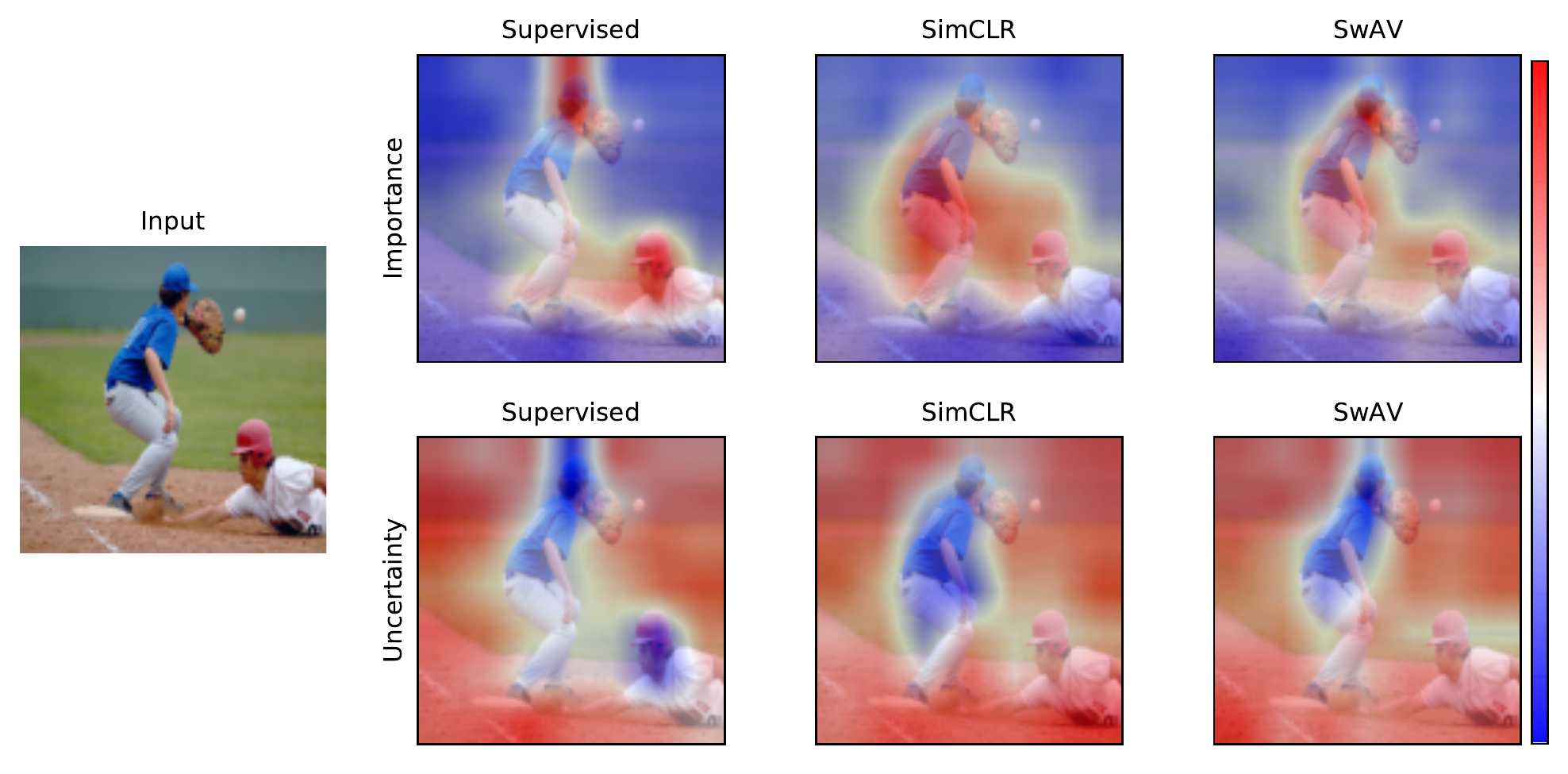}
    \caption{Example from the COCO dataset.}
    \label{fig:ex4}
\end{figure*}

\begin{figure*}[htb]
    \centering
    \includegraphics[width=0.75\linewidth]{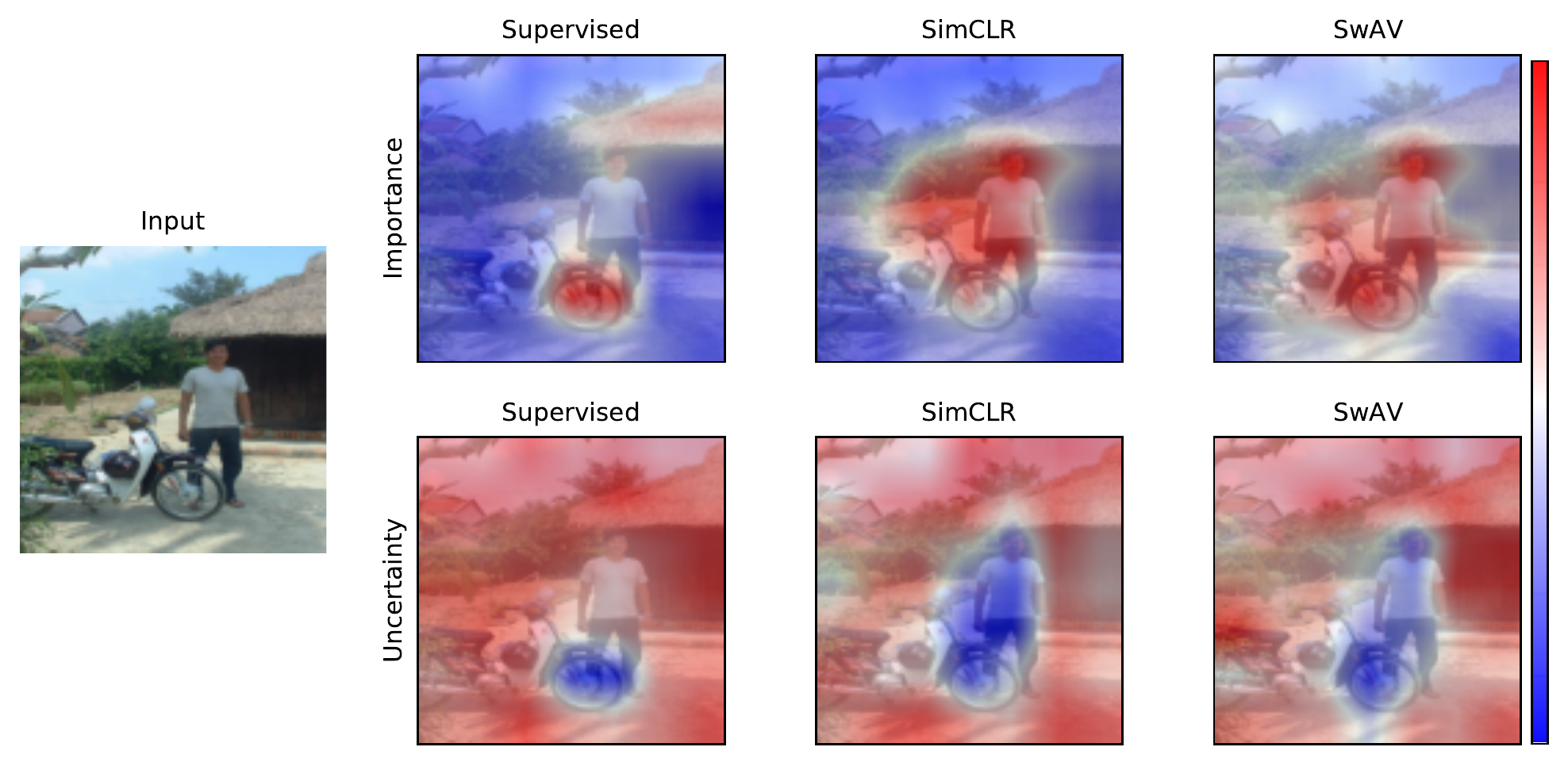}
    \caption{Example from the VOC dataset.}
    \label{fig:ex5}
\end{figure*}

\begin{figure*}[htb]
    \centering
    \includegraphics[width=0.75\linewidth]{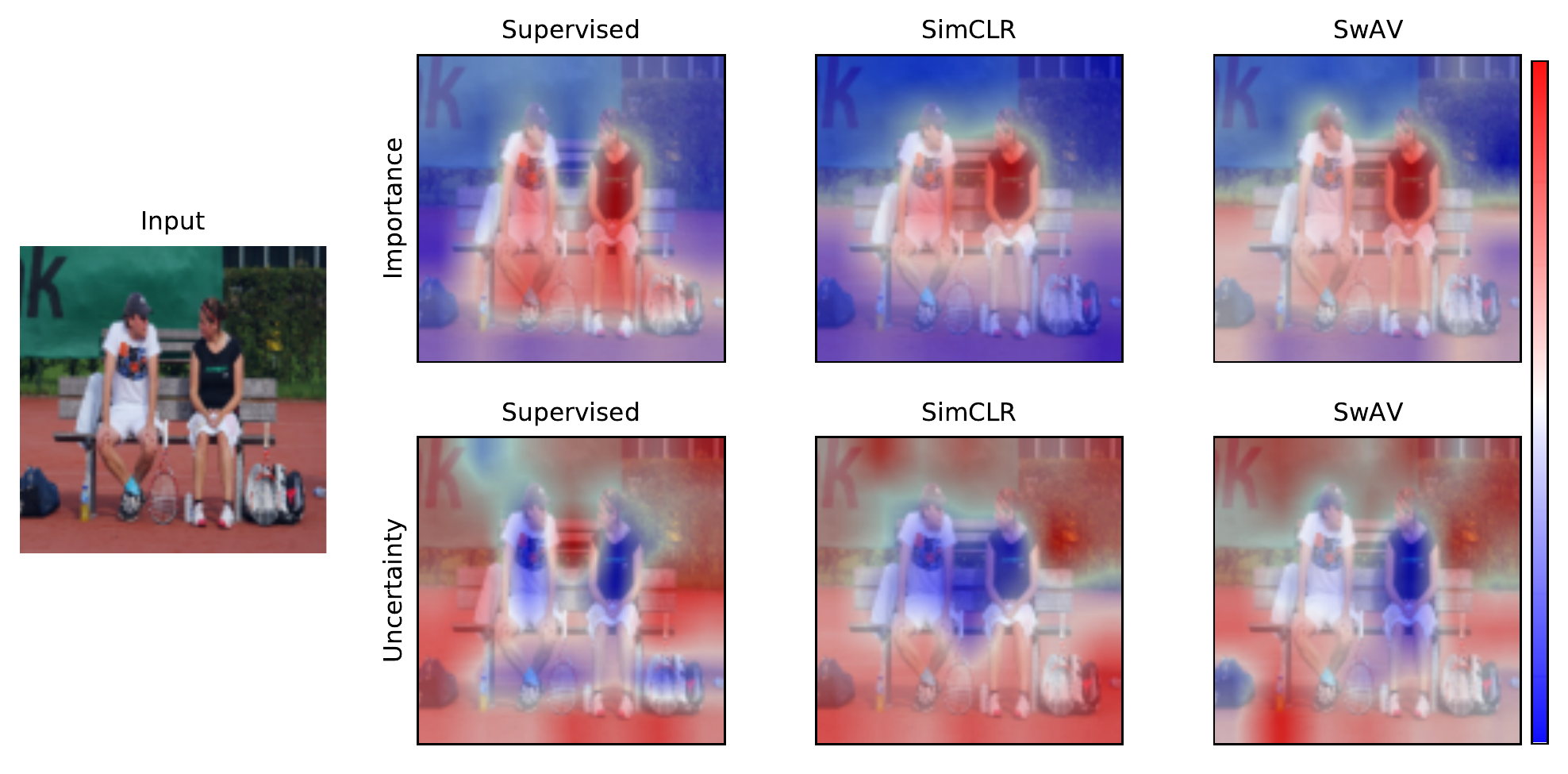}
    \caption{Example from the COCO dataset.}
    \label{fig:ex6}
\end{figure*}

\begin{figure*}[htb]
    \centering
    \includegraphics[width=0.75\linewidth]{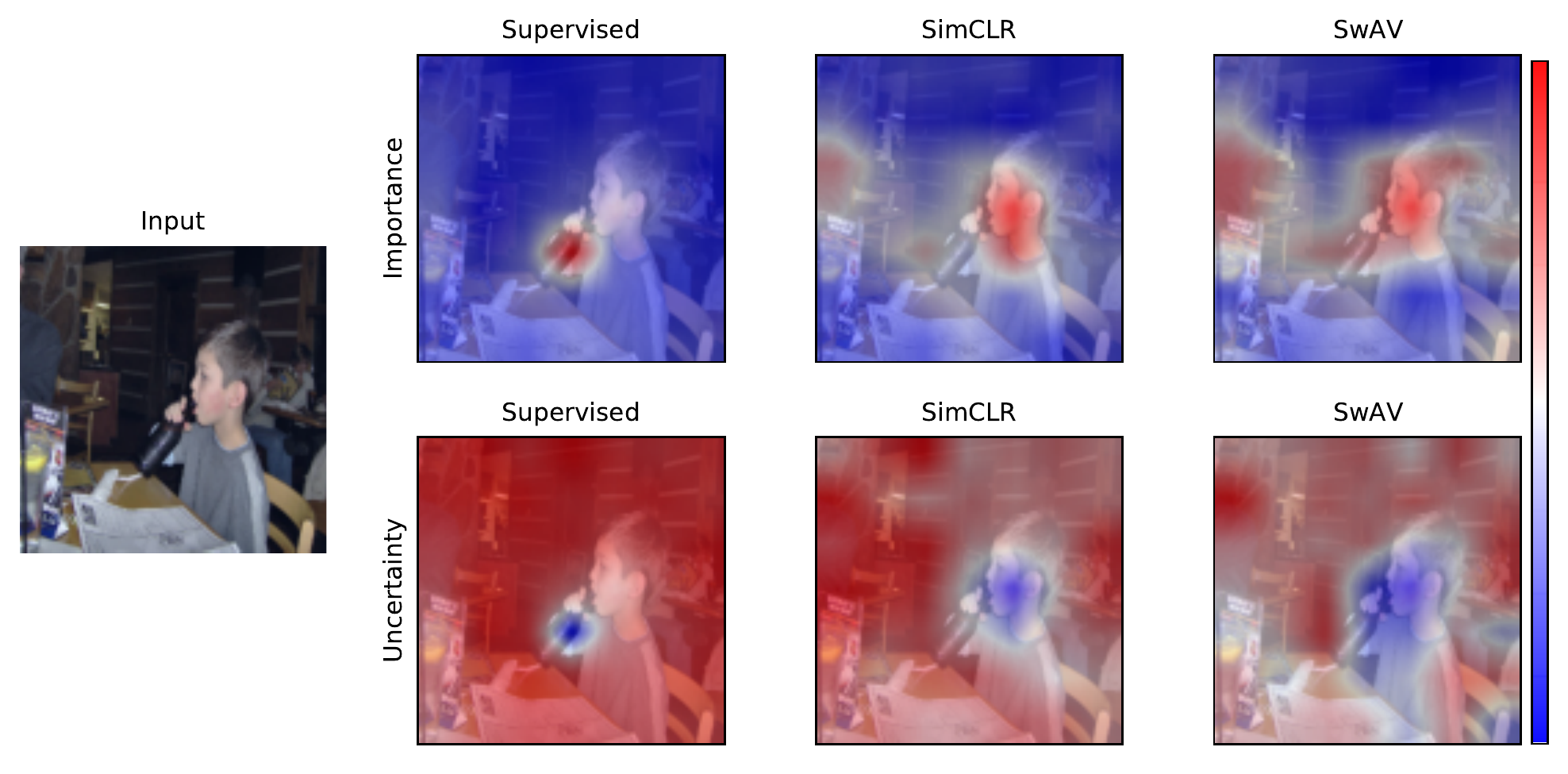}
    \caption{Example from the VOC dataset.}
    \label{fig:ex7}
\end{figure*}

\begin{figure*}[htb]
    \centering
    \includegraphics[width=0.75\linewidth]{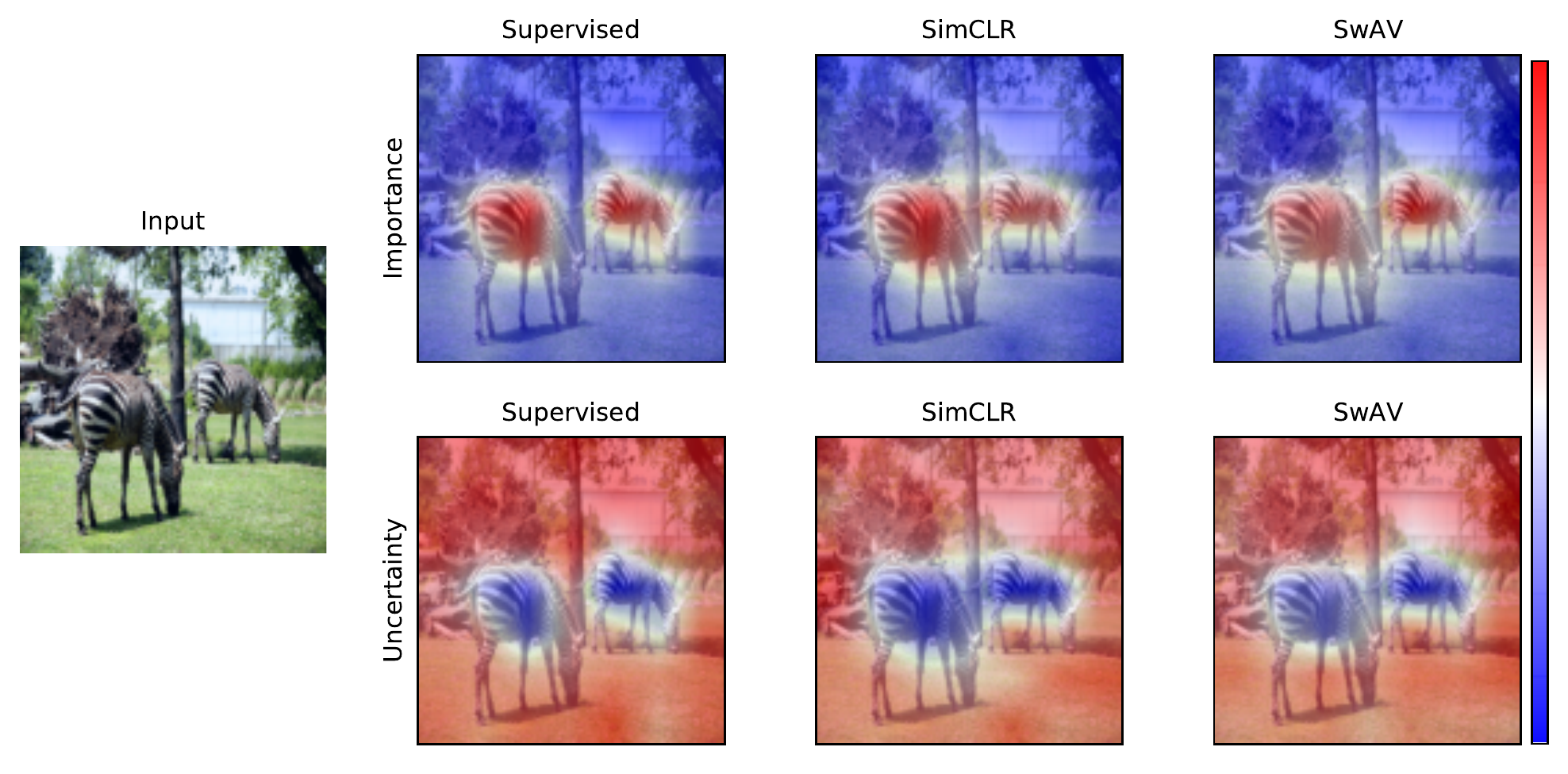}
    \caption{Example from the COCO dataset.}
    \label{fig:ex8}
\end{figure*}

\begin{figure*}[htb]
    \centering
    \includegraphics[width=0.75\linewidth]{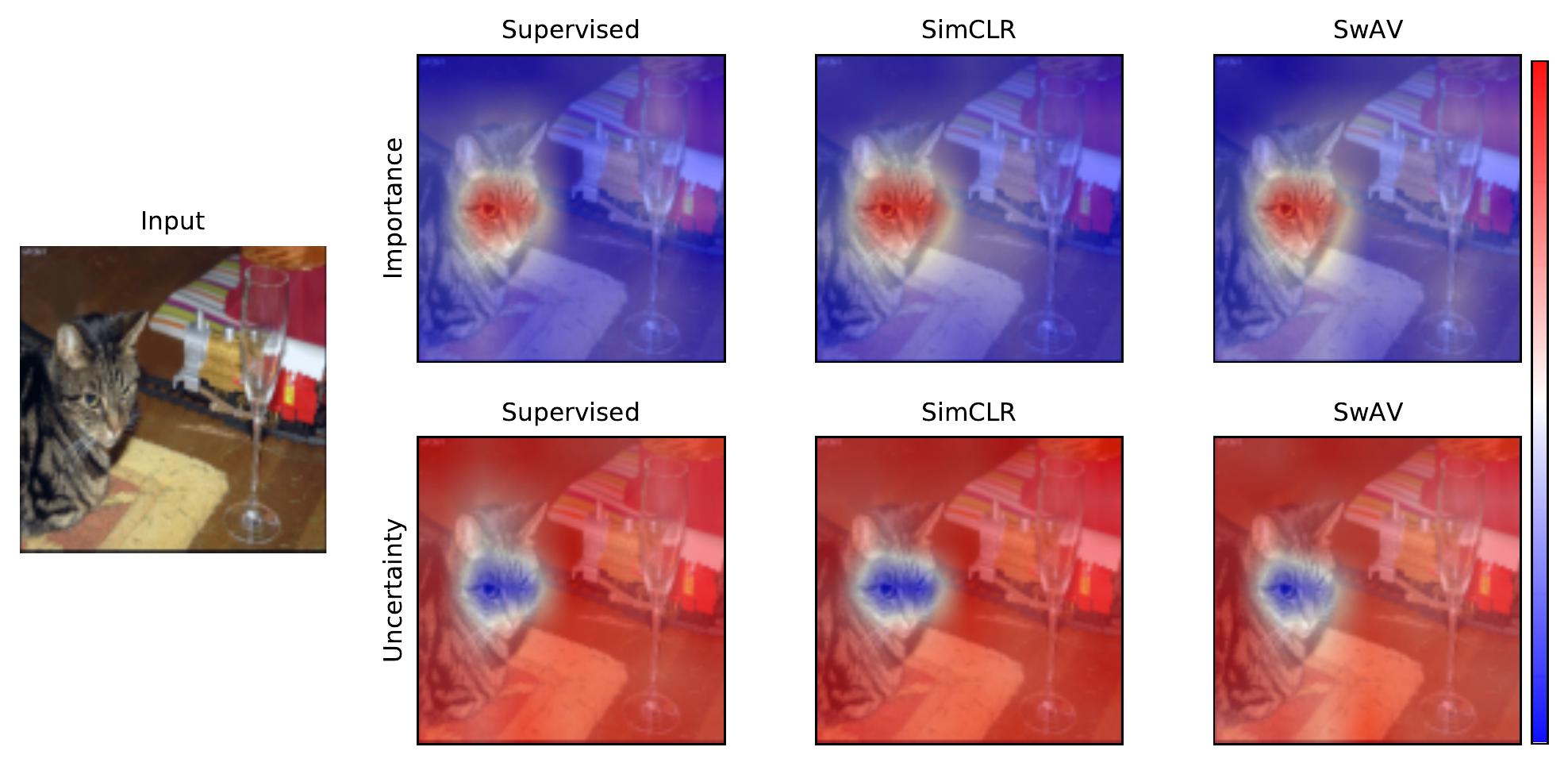}
    \caption{Example from the VOC dataset.}
    \label{fig:ex9}
\end{figure*}

\begin{figure*}[htb]
    \centering
    \includegraphics[width=0.75\linewidth]{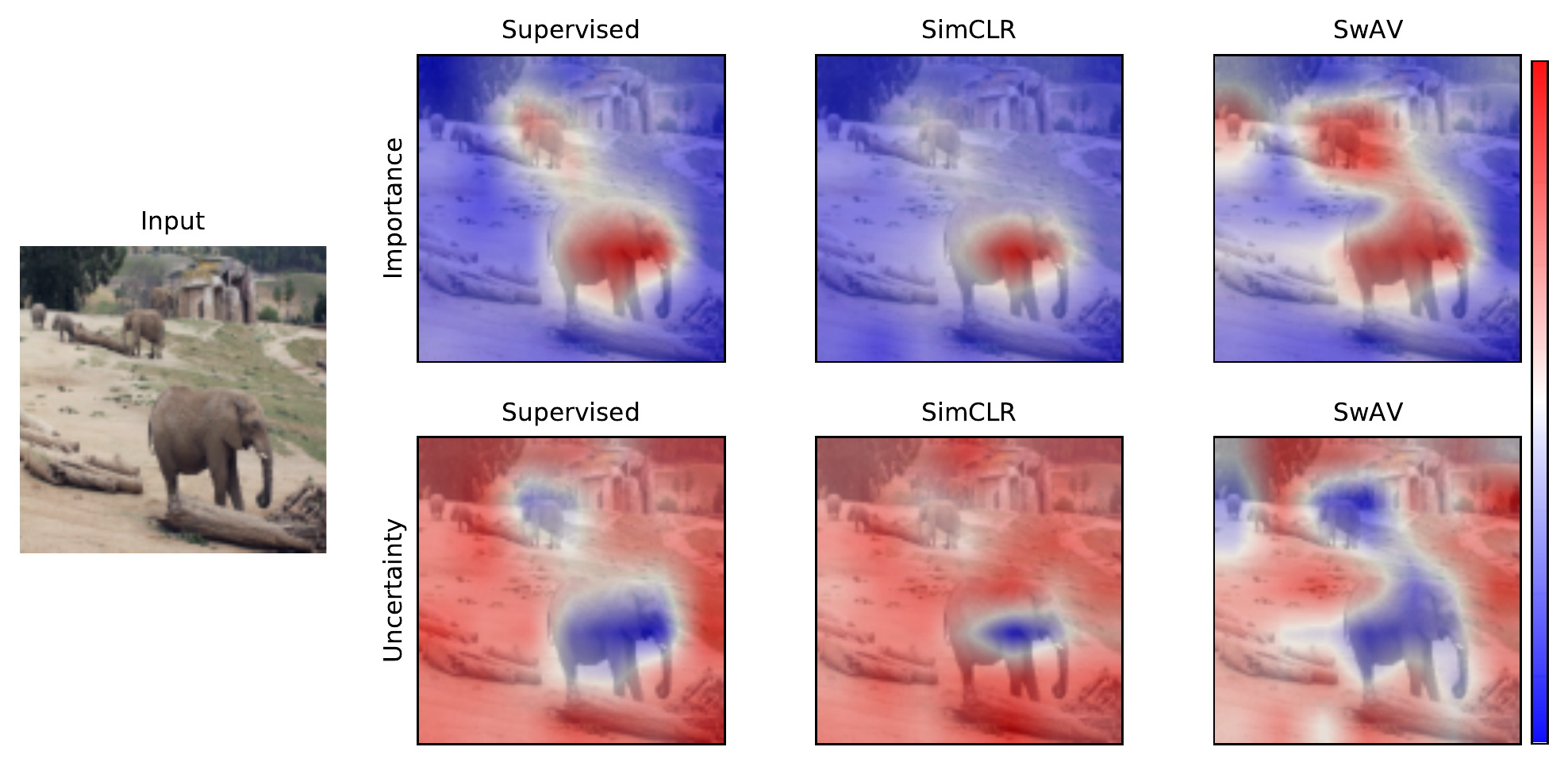}
    \caption{Example from the COCO dataset.}
    \label{fig:ex10}
\end{figure*}

This section presents additional qualitative results. Figure \ref{fig:ex1} to \ref{fig:ex10} displays examples of explanations and their associated uncertainty, provided by RELAX, for images from the VOC and COCO dataset. Figure \ref{fig:ex1} displays an example where all feature extractors agree in terms of importance, but the degree of uncertainty varies. Figure \ref{fig:ex2} shows an example where only SwAV highlight both objects as important for the representation. Similarly, Figure \ref{fig:ex3} displays an example where only SwAV is considering both the person and the car as important for the representation. Figure \ref{fig:ex3} to \ref{fig:ex10} shows similar examples where RELAX provides insights into the different feature extractors.

\end{appendices}


\clearpage
\clearpage
\bibliography{bibliography}


\end{document}